 \documentclass[final,5p,times,twocolumn]{elsarticle}

\usepackage[english]{babel}
\usepackage{lineno}
\usepackage{amssymb, amsmath}
\usepackage{amsmath,amsfonts,amssymb,amsthm,epsfig,url,epstopdf,array}
\usepackage{graphicx}
\usepackage[justification=centering]{caption}
\usepackage{multirow}
\usepackage{hyperref}
\usepackage{enumerate}
\usepackage{color}
\usepackage[pdftex,dvipsnames]{xcolor}
\usepackage{float}
\usepackage{subcaption}
\usepackage[flushleft]{threeparttable}
\usepackage{booktabs,caption}
\usepackage{xargs}
\usepackage{pgf}
\usepackage{tikz}
\usepackage{relsize}
\usetikzlibrary{arrows,automata,positioning}
\usepackage[flushleft]{threeparttable} 
\usepackage{algorithm}
\usepackage{algpseudocode}
\usepackage{color, colortbl}
\definecolor{Gray}{gray}{0.9}

\hypersetup{pdfauthor={Name}}

\newtheorem{theorem}{Theorem}

\newtheorem{corollary}{Corollary}[theorem]

\journal{Neurocomputing}

\begin{document}
\begin{frontmatter}

\title{Modeling Implicit Bias with Fuzzy Cognitive Maps}


\author[TILBURG]{Gonzalo N\'apoles\corref{mycorrespondingauthor}}
\address[TILBURG]{Department of Cognitive Science \& Artificial Intelligence, Tilburg University, The Netherlands.}

\cortext[mycorrespondingauthor]{Corresponding author}
\ead{g.r.napoles@uvt.nl}

\author[TUe]{Isel Grau}
\address[TUe]{Information Systems Group, Eindhoven University of Technology, The Netherlands.}

\author[HASSELT,UCLV]{Leonardo Concepci\'on}
\address[HASSELT]{Business Informatics Research Group, Hasselt University, Belgium.}
\address[UCLV]{Department of Computer Science, Central University of Las Villas, Cuba.}

\author[HASSELT]{Lisa Koutsoviti Koumeri}

\author[BRASIL]{João Paulo Papa}
\address[BRASIL]{Department of Computing, São Paulo State University, Brazil}

\begin{abstract}
This paper presents a Fuzzy Cognitive Map model to quantify implicit bias in structured datasets where features can be numeric or discrete. In our proposal, problem features are mapped to neural concepts that are initially activated by experts when running what-if simulations, whereas weights connecting the neural concepts represent absolute correlation/association patterns between features. In addition, we introduce a new reasoning mechanism equipped with a normalization-like transfer function that prevents neurons from saturating. Another advantage of this new reasoning mechanism is that it can easily be controlled by regulating nonlinearity when updating neurons' activation values in each iteration. Finally, we study the convergence of our model and derive analytical conditions concerning the existence and unicity of fixed-point attractors.
\end{abstract}


\end{frontmatter}


\section{Introduction}
\label{sec:introduction}

Automated or algorithmic decision-making is nowadays used by organizations worldwide partly because machines seem neutral and unbiased~\cite{zuiderveen2020strengthening, vzliobaite2017measuring}. Such an assumption is rather naive since bias can be encoded into historical data by capturing the discriminatory beliefs of people involved in the data generation process~\cite{mehrabi2021survey}. Artificial Intelligence (AI) algorithms then use the contaminated data to facilitate life-changing decisions such as accepting loan applications or suggesting appropriate medical treatments. Therefore, detecting bias in data used by Machine Learning (ML) systems is of paramount importance to the welfare of both society and individuals~\cite{hajian2012methodology}. 

Bias in decision-making tasks can be expressed implicitly or explicitly~\cite{mehrabi2021survey}. Direct discrimination (or explicit bias) occurs when decisions are influenced by sensitive or protected features like gender, race, or marital status. Indirect discrimination (or implicit bias) occurs when decisions are influenced by non-sensitive features that strongly correlate with sensitive ones, thus resulting in non-favorable outcomes towards underprivileged groups. Implicit bias is also referred to in the literature as unconscious bias~\cite{hoffmann2019fairness}. It expresses unintentional forms of discrimination that infect decision-making structurally and systematically and remains hard to address during data collection, labeling, or at an algorithmic level~\cite{hoffmann2019fairness}. 

Available literature dedicated to discovering implicit bias is scarce compared to respective literature regarding explicit bias mainly because the former is harder to uncover~\cite{vzliobaite2017measuring, ruggieri2014anti}. We identify three main paradigms to discover and measure indirect discrimination: (i) statistical tests and correlation~\cite{vzliobaite2017measuring}, (ii) causal models~\cite{makhlouf2020survey}, and (iii) classification rules~\cite{hajian2012methodology}. We sum up the limitations of these approaches that are relevant to our proposed model. Firstly, several methods require discretization of numeric features, which might differ per case and/or distort the information encoded in the data. Secondly, some approaches are still dependent on black-box machine learning models whose outputs are dependent on data pre-processing or the train-test split. Thirdly, all approaches consider interactions among features or feature category pairs selected by experts. However, experts might misjudge the impact of feature categories on the decision outcomes~\citep{grgic2018beyond}. Finally, all approaches do not account for every possible interaction among the rest of the concepts within the system.

In previous work, we proposed a bias quantification measure based on granular computing, called fuzzy-rough uncertainty (FRU), for quantifying explicit bias ~\cite{koutsoviti2021bias}. FRU relies on fuzzy-rough sets to measure the extent to which the exclusion of protected features from a structured dataset changes the boundary regions. Positive changes mean that the boundary regions expanded themselves, which can be considered a proxy for explicit bias. FRU values measured after removing protected features from the data are not absolute and need to be interpreted concerning the respective FRU values of unprotected features. Extending this concept to detect implicit bias in~\cite{Napoles2021Afuzzyrough}, we hypothesize that a strong correlation\footnote{~For simplicity purposes, the term correlation simultaneously represents three different measures chosen based on variable type (numeric or nominal) described in Section~\ref{sec:proposal:model:correlation}: Pearson correlation, strength of association and R-squared coefficient of determination.} between a protected feature and unprotected ones coupled with a low FRU value of the protected feature in question can reveal which unprotected features might be (to some extent) implicitly biased.

Table~\ref{tab:correlation_interpretation} shows the possible scenarios concerning implicit and explicit bias, as discussed in~\cite{Napoles2021Afuzzyrough}. A limitation of such a four-scenario approach is that one would need to analyze all possible pairwise feature combinations to study the implicit bias patterns in the data, and their interpretation might be too complex. Moreover, such a pairwise analysis would not suffice when bias is implicitly encoded by more than two features simultaneously or arises from feedback loops. To address this limitation, we propose a solution that considers interactions among all combinations of variables simultaneously. 

\begin{table}[!ht]
\centering
\caption{Scenarios relating correlation and FRU values.}
\resizebox{0.99\linewidth}{!}{
\begin{tabular}{|p{3.5cm}|p{3.5cm}|p{3.5cm}|}
\hline
{} & \cellcolor{Gray} Large FRU value & \cellcolor{Gray} Small FRU value \\ 
\hline
\cellcolor{Gray} Strong correlation & Explicit \& Implicit bias & Implicit bias \\ 
\hline
\cellcolor{Gray} Weak correlation & Explicit bias & Safe scenario\\
\hline
\end{tabular}}
\label{tab:correlation_interpretation}
\end{table}


In this paper, we introduce a model based on Fuzzy Cognitive Maps (FCMs)~\cite{Kosko1986} to quantify the amount of bias towards protected features given initial conditions related to unprotected ones. In our FCM model, neural concepts denote protected or unprotected problem features, and weights comprise correlation patterns encoding the implicit bias. To run what-if simulations with this model, we activate neurons denoting (selected) unprotected features, execute the reasoning mechanism, and measure the amount of bias collected by neurons denoting protected features. Other contributions of our research concern theoretical challenges about FCM's reasoning and convergence that have remained unsolved until now. Firstly, we couple the quasi nonlinear reasoning rule proposed in \cite{Napoles2021} with a normalization-like transfer function that promotes competition among the neurons in each iteration. Such a quasi nonlinear reasoning rule involves a parameter controlling the nonlinearity degree of the model. Secondly, we analytically study the convergence properties of our model when it comes to the existence and uniqueness of the fixed point. Overall, we derive conditions guaranteeing (i) that the fixed point exists and is unique, or (ii) that the fixed point cannot be unique. The former is useful when modeling invariant situations that do not depend on the initial conditions, while the latter is useful when running what-if simulations. Section~\ref{sec:fcm} further motivates these theoretical contributions after having introduced the FCM formalism.

The rest of this paper is organized as follows. Section~\ref{sec:literature} revisits the literature devoted to detecting and quantifying indirect (implicit) bias in structured datasets. Section~\ref{sec:fcm} presents the theoretical background concerning fuzzy cognitive modeling. Section~\ref{sec:proposal} introduces the proposed FCM-based model and the new reasoning mechanism, while Section~\ref{sec:convergence:} elaborates on the convergence properties of our reasoning mechanism and derives analytical conditions related to fixed-point attractors. Section \ref{sec:simulations}~presents the simulation results using a well-known case study, and Section \ref{sec:remarks} states concluding remarks.

\section{Related Works on Bias Detection and Quantification}
\label{sec:literature}

In this section, we briefly introduce the main paradigms used to discover indirect discrimination (or implicit bias) in the literature and mention their limitations. 

The most popular approach to discover implicit bias comes from the field of statistics using regression models or correlation coefficients~\cite{vzliobaite2017measuring, tramer2017fairtest}. However, such methods are not easily generalizable~\cite{vzliobaite2017measuring}, and higher-order interactions between many features and the target are hard to interpret. The authors in~\cite{bonchi2017exposing} also mention that state-of-the-art methods are mainly correlation-based while stressing out that correlation could be spurious and does not necessarily imply causation. On the one hand, statistical methods can be translated to algorithmic discrimination measures and optimization constraints~\cite{vzliobaite2017measuring}. On the other hand, one might wonder whether modeling implicit bias must have causal semantics.

A second approach to discover implicit bias relies on generating classification rules learned from the data as reported in ~\cite{hajian2012methodology}. In such a work, the authors use predetermined discriminatory items (feature-category pairs) and the concepts of support and confidence of classification rules to quantify and correct implicit and explicit bias. Their measure computes the correlation between predefined and non-discriminatory items to identify redlining or biased rules. This approach has two potential limitations. First, the decision variable and protected features must be binary or nominal, requiring numeric features to be discretized. Second, manually predetermining the sensitive feature categories might be risky as human experts might misjudge their impact on the decision outcomes~\cite{grgic2018beyond}.

The authors in~\cite{adler2018auditing} introduce a method to quantify indirect influence using black-box prediction models. First, they estimate the information content of a feature by predicting it from the remaining features. Then, they modify or obscure the feature so that it can no longer be predicted from the data and, finally, analyze the difference between the original feature and the modified one. Possible limitations of this approach are their measure's dependency on the data distribution, the need to discretize numeric attributes, and the black-box manner in which predictions are made. The authors applied their suggested method to the German Credit dataset~\cite{Dua:2019} and concluded that implicit bias is evident in the features \textit{credit amount}, \textit{checking status} and \textit{existing credits}.

A third approach to discover implicit bias is based on causal networks~\cite{makhlouf2020survey}. Causal models assume that implicit bias is transmitted as causal effects along paths that pass through proxy attributes~\cite{zhang2018causal}. Zhang et al.~\cite{zhang2018causal} attempt to discover implicit bias in historical data by coupling causal networks and probabilistic inference. They use the path-specific effects as a proxy to model bias. The authors in~\cite{bonchi2017exposing} also use probabilistic causal theory to detect implicit bias using the German Credit dataset. They examine the existence of implicit bias between three protected features (\textit{gender}, \textit{age} and \textit{foreign worker}) and five unprotected features (\textit{employed since} being the only one that is also considered in our approach), but report that discrimination patterns are less palpable partly because this dataset contains less proper causal relations. Their conclusion states that automatically exploring the whole space of possible discrimination patterns is an ongoing research challenge. 

Causal paths are also explored by Chiappa in ~\cite{chiappa2019path} combined with the notion of counterfactual fairness in order to compute indirect causal effects. In order to eliminate unfair information, the authors apply a correction that is suitable to complex nonlinear models, and its main limitation is that the data-generation process needs to be provided. In their numerical simulations using the German Credit dataset, they conclude that there is evidence of implicit bias between the features \textit{housing} and \textit{gender} which aligns with our findings. 

In conclusion, the causal approach considers one-way interactions among a limited number of features or feature categories, whereas an FCM-based approach using a square correlation matrix as weights accounts for interactions among all features simultaneously. This is also the case with the classification rule-based approach. Moreover, both approaches use the target variable during the process of measuring indirect bias in contrast to our FCM-based approach, where information flow among variables is modeled without being affected by the possibly partial decision feature. 

\section{Fuzzy Cognitive Maps}
\label{sec:fcm}

FCMs were introduced in~\cite{Kosko1986} as a knowledge-based method for modeling complex systems and running what-if simulations. Such interpretable recurrent neural networks consist of neural concepts and weighted connections. Neural concepts represent variables, entities, or states related to the physical system under investigation. The signed weight associated with each connection denotes the strength of the causality (or simply correlation) between the corresponding variables. Weights are quantified in the $[-1,1]$ interval, while the neurons' activation values can take values in either $[0,1]$ or $[-1, 1]$ depending on the nonlinear transfer function attached to each neural unit.

In each iteration, FCM-based models update the activation values of neural concepts based on the neurons' states in the previous iteration and the weight matrix. Overall, the recurrent reasoning process allows several times to propagate a given initial condition across the whole network, which might be necessary to model scenarios with feedback loops.

Equation \eqref{eq:reasoning-old} shows the reasoning rule to compute the activation value of the $i$-th neural concept in the $(t+1)$-th iteration for a given initial condition provided by the domain expert when running what-if simulations: 

\begin{equation}
a_i^{(t+1)} = f \left(\sum_{j=1}^M a_j^{(t)} w_{ij} \right),
\label{eq:reasoning-old}
\end{equation}

\noindent where $M$ is the number of neurons and $w_{ij}$ is the weight connecting the $C_i$ and $C_j$ neurons, while $f(\cdot)$ is the transfer function used to keep the neurons' activation values within the allowed activation interval. The sigmoid function below is often used as a transfer function in neural systems: 
\begin{equation}
f(x) = \frac{1}{1+e^{-\lambda x}},
\label{eq:sigmoid}
\end{equation}

\noindent such that $\lambda > 0$ is the inclination parameter to control the function steepness. Larger values of $\lambda$ cause the sigmoid function to resemble a binary activator. It is worth mentioning that we can use other bounded, continuous transfer functions such as the hyperbolic tangent function, but we need to re-evaluate how to interpret the simulation results.

The reasoning rule stops when either (i) the model converges to a fixed point or (ii) a maximal number of iterations $T$ is reached. Overall, we have three possible states:

\begin{itemize}
  \item \textbf{Fixed point} $(\exists t_{\alpha} \in \{1,\dots,(T-1)\} : a_i^{(t+1)}=a_i^{(t)}, \forall i, \forall t \geq t_{\alpha})$: the FCM produces the same state vector after $t_{\alpha}$, thus $a_i^{(t_{\alpha})}=a_i^{(t_{\alpha}+1)}=a_i^{(t_{\alpha}+2)}=\dots=a_i^{(T)}$. If the fixed point is unique, the FCM model will produce the same state vector regardless of the initial conditions. The unique fixed-point attractor allows modeling invariant situations rather than performing what-if simulations.
  \item \textbf{Limit cycle} $(\exists t_{\alpha},P,j \in \{1,\dots,(T-1)\}: a_i^{(t+P)} = a_i^{(t)},$ $ \forall i, \forall t \geq t_{\alpha})$: the FCM produces the same state vector periodically after the period $P$, thus $a_i^{(t_{\alpha})}=a_i^{(t_{\alpha}+P)}=a_i^{(t_{\alpha}+2P)}=\dots=a_i^{(t_{\alpha}+jP)}$, where $t_{\alpha}+jP \leq T$.
  \item \textbf{Chaos}: the FCM produces different state vectors for successive iterations with no clear pattern.
\end{itemize}

The convergence of FCM-based models is of paramount importance when running what-if simulations. However, we have noticed that most published papers using FCMs to model complex systems rarely touch upon the practical implications of having multiple or unique fixed points, or cyclic patterns. There could be several reasons for such a gap. Firstly, the nonlinearity of FCM-based models operating as closed systems makes them difficult to control. Secondly, the most prominent theoretical results reported in the literature~\cite{Boutalis2009,Harmati2020,Knight2014} are primarily devoted to deriving sufficient conditions ensuring the existence and uniqueness of the fixed point. These conditions might be quite strict and difficult to fulfill in reality. Moreover, they have limited usability if we want to perform what-if simulations where the outputs should not remain static.

Another issue that emerges when using bounded transfer functions and large, densely connected FCM-based models is that the neuron's activation values often locate either in the lower or upper boundary of the activation interval. Hence, the neurons tend to \textit{saturate} when receiving a large negative or positive information flow. One way to address this drawback is to modify the reasoning rule to promote competition among the neurons in each iteration. Finally, some transfer functions such as the sigmoid function can distort the simulation results by activating neurons that are not supposed to be active given the information flow they receive.

In the following subsection, we overcome these drawbacks and propose a sound FCM-based model to perform simulations concerning the impact of the implicit bias over protected features in pattern classification problems. 

\section{Modeling implicit bias using cognitive modeling}
\label{sec:proposal}

This section presents our FCM-based simulation model to quantify the amount of implicit bias towards protected features given initial conditions imposed by (selected) unprotected features. Firstly, we further discuss the problem of quantifying implicit bias and introduce the intuition of our model. Secondly, we describe how to build the FCM structure using the correlation/association patterns between problem features. Thirdly, we introduce a re-scaled transfer function coupled with a parametric quasi nonlinear reasoning rule to prevent saturation and convergence issues from happening.

\subsection{Motivating our FCM-based model}
\label{sec:proposal:model:intuition}

One of the difficulties in quantifying implicit bias is that protected features can be affected by seemingly unrelated features through hidden patterns. For example, the postal code might influence whether or not an applicant gets a loan because people living in more expensive neighborhoods might be more likely to receive the loan. Such hidden patterns result from complex dependencies, correlations, associations, causalities, and feedback loops that are difficult to detect by inspecting the data. This suggests that implicit bias should be modeled as a \textit{complex system} involving feedback loops and non-trivial dependencies. Note that having feedback loops justifies using recurrent neural networks even when the target problem does not involve sequences or an explicit time component.


In our paper, modeling implicit bias means building an FCM-based model to quantify the bias amount for protected features given initial conditions associated with unprotected ones. The model will enable performing what-if simulations with the form: ``What is the amount of bias attached to the protected feature $y \in Y$ if $x_1$ is $a_1^{(0)}$, $x_2$  is $a_2^{(0)}$, $\ldots$, and $x_n$ is $a_n^{(0)}$, such that $Y$ is the set of protected features and $X$ is the set of unprotected ones. The initial conditions $a_1^{(0)}$, $a_2^{(0)}$, $\ldots$, $a_n^{(0)}$ should be determined by domain experts when running the simulations. To perform such simulations, we need to ensure that the fixed point is not unique. At the same time, we might want to quantify the amount of bias towards a protected feature regardless of these initial conditions, just relying on the weight matrix. By doing so, we need to ensure that the fixed point is unique to prevent that the FCM-based model converges to a different fixed point as the initial conditions change.

\subsection{Building the correlation network}
\label{sec:proposal:model:correlation}

The first step towards modeling implicit bias using cognitive mapping is to determine the network structure (that is to say, the concepts and the weight matrix). Each problem feature is denoted with a neural concept in our neural reasoning system regardless of whether the feature is nominal or continuous. The initial activation values of these neurons represent the extent to which the corresponding feature is active in the model. Initial activation values are defined by the experts when running what-if simulations. In contrast, the neurons' activation values resulting from the recurrent reasoning process (see the following subsection) represent the amount of implicit bias given the initial condition provided by the domain expert.

The weights connecting the neurons represent correlation patterns to be derived from the dataset being analyzed. The intuition behind these weights is that if a protected feature correlates with an unprotected one, then part of the bias will be encoded in the unprotected feature. Hence, capturing the information that arrives at a neuron representing a protected feature given an initial condition is reliable for detecting implicit bias. More importantly, our approach detects implicit bias considering non-trivial relationships since the interaction among variables is modeled as a complex system.

The weight matrix is obtained using three measures that are conceptually similar (i.e. based on Pearson's correlation coefficient) in an effort to preserve consistency. Let $C_i$ and $C_j$ be two neural concepts in the network. If both $C_i$ and $C_j$ represent numeric features, then $w_{ij}$ will be given by the Pearson correlation coefficient~\cite{rovine199714th} (even though features in German Credit dataset do not meet the assumptions of normality, linearity or homoscedasticity). If both $C_i$ and $C_j$ represent nominal features, then $w_{ij}$ will denote the association strength between these variables as computed by the Cram\'er's V~\cite{cramer2016mathematical} which is based on Pearson’s chi-squared statistic~\cite{berry2018measurement}. If $C_i$ represents a nominal feature and $C_j$ represents a numeric feature, then $w_{ij}$ is the R-squared coefficient of determination~\cite{nagelkerke1991note} which is the square of the Pearson’s correlation. This measure quantifies the percentage of variation in the numeric feature that is explained by the nominal one coupled with an F-test of joint significance~\cite{kramer2005r2}. Notice that the weight matrix is symmetric such that $w_{ij}=w_{ji}$. Finally, we compute the absolute values of these weights since we are not interested in the correlation signs but their intensity. It is important to state that in this paper, the term correlation simultaneously represents all the three measures described above for simplicity.

\subsection{Quasi nonlinear reasoning rule}
\label{sec:proposal:model:quasinonlinear}

As mentioned earlier, the reasoning rule in Equation \eqref{eq:reasoning-old} often converges to unique fixed-point attractors or suffers from saturation problems. The former could be useful when modeling invariant situations, but it is less attractive when performing what-if simulations with some generalization. For example, if we want to design an FCM-based model to solve a regression-like problem and the model converges to a unique fixed-point attractor, the predicted value will be the same regardless of the inputs. Moreover, when having dense FCM-based models, neurons often produce saturated activation values even when their inputs are significantly different. In this paper, we couple the quasi nonlinear reasoning rule recently proposed in~\cite{Napoles2021} with a re-scaled transfer function that performs a normalization operation. The resulting reasoning model solves the issues mentioned above while producing relative activation values, thus making the model more intuitive and easy to control.

Let $\mathbf{A}^{(t)}=( a_{1}^{(t)}, \ldots, a_{i}^{(t)}, \ldots, a_{M}^{(t)} )$ be the activation vector produced by an FCM such that $a_{i}^{(t)}$ is the activation value of the $i$-th neuron at the iteration $t$. Similarly, let $\bar{\mathbf{A}}^{(t)}=(\bar{a}_{1}^{(t)}, $ $\ldots, \bar{a}_{i}^{(t)}, $ $\ldots, \bar{a}_{M}^{(t)})$ be the raw activation vector where $\bar{a}_{i}^{(t)}$ is the raw activation value of the $i$-th neuron in the current iteration. In other words, the vector $\bar{\mathbf{A}}^{(t)}$ is the argument of the transfer function and can be calculated as $\bar{\mathbf{A}}^{(t)} = \mathbf{A}^{(t)}\mathbf{W}$ where $\mathbf{W}_{M \times M}$ is the weight matrix defining the interaction among the neurons. The quasi nonlinear reasoning rule using the re-scaled transfer function can be computed as follows:

\begin{equation}
a_i^{(t+1)} = \underbrace{\phi f \left( \sum_{j=1}^M a_j^{(t)} w_{ij} \right)}_{\text{nonlinear component}} + \underbrace{\left( 1-\phi \right)a_i^{(0)}}_{\text{linear component}},
\label{eq:reasoning-new}
\end{equation}
\noindent where

\begin{equation}
f\left(\bar{a}_{i}^{(t+1)}\right) = \frac{\bar{a}_{i}^{(t+1)}}{\left\Vert \bar{\mathbf{A}}^{(t+1)} \right\Vert_2},
\label{eq:re-scaled}
\end{equation}

\noindent such that $||\cdot||_2$ stands for the Euclidean norm, and $0 \leq \phi \leq 1$ controls the nonlinearity of the reasoning rule. When $\phi=1$, the model performs as a closed system where the activation value of a neuron depends on the activation values of connected neurons in the previous iteration. When $0<\phi<1$, we add a linear component to the reasoning rule devoted to preserving the initial activation values of neurons when updating their activation values in the current iteration. When $\phi=0$, the model narrows down to a linear regression where the initial activation values of neurons act as regressors. It should be mentioned that, when the denominator of the nonlinear component is zero and $\phi \neq 0$, then the neuron's activation value will be given by $\left( 1-\phi \right)a_i^{(0)}$. Two reasons move this rule: (i) we need to avoid indetermination issues in the denominator, and (ii) we want to prevent situations in which the neurons receiving no information flow at all get arbitrarily activated by the transfer function. For example, the sigmoid transfer function produces $0.5$ when its argument is zero, leading to misleading simulation results.

\begin{figure*}[!htbp]
\center
	\begin{subfigure}{0.33\textwidth}
	\center
	\includegraphics[width=\textwidth]{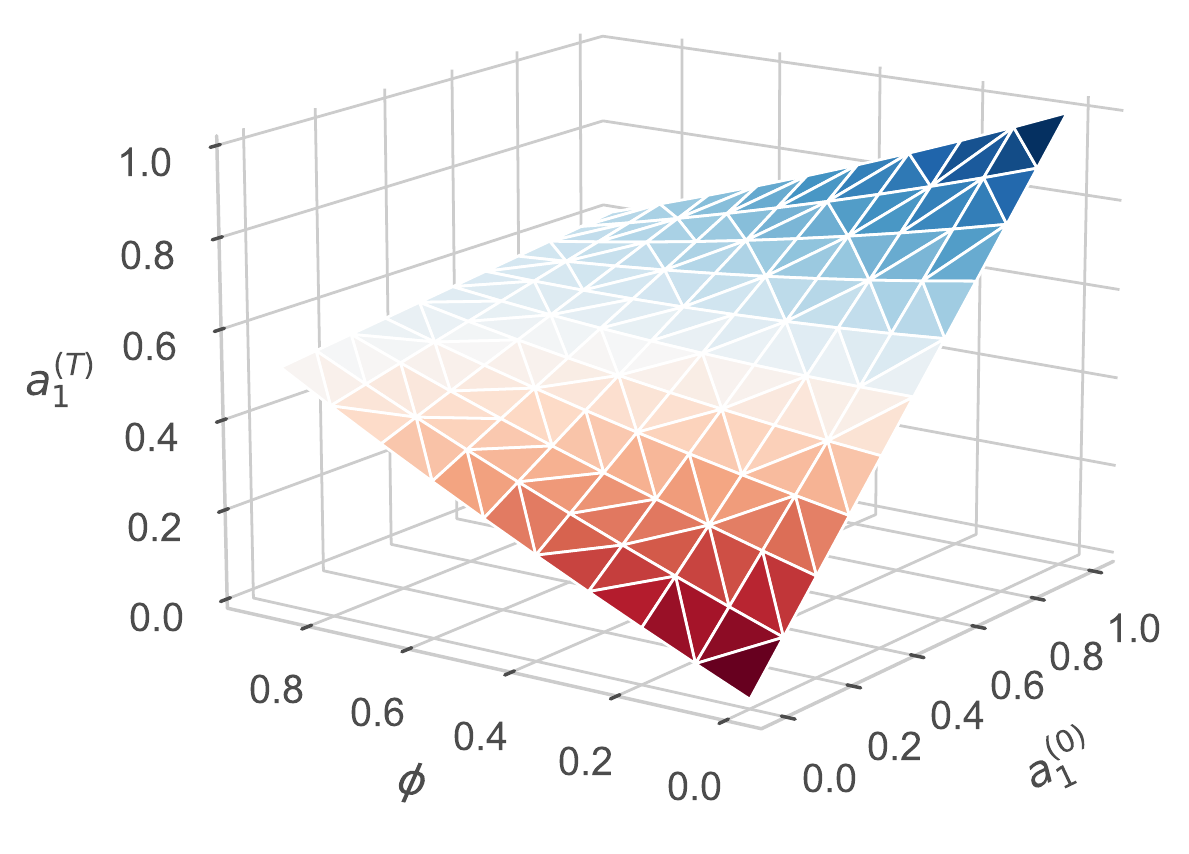}
	\caption{Sigmoid activation function}
	\label{fig:act_sigmoid}
	\end{subfigure}
	\begin{subfigure}{0.33\textwidth}
	\center
	\includegraphics[width=\textwidth]{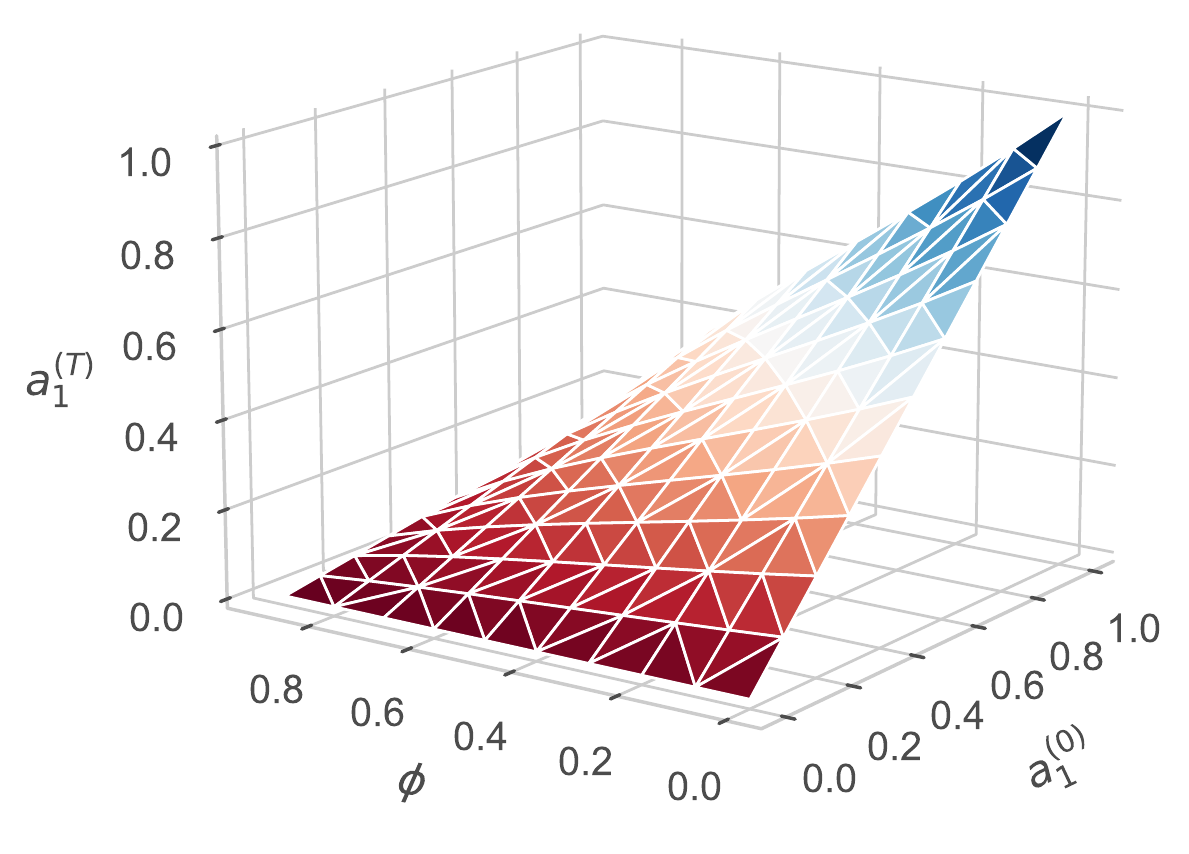}
	\caption{Hyperbolic activation function}
	\label{fig:act_hyperbolic}
	\end{subfigure}
    \begin{subfigure}{0.33\textwidth}
	\center
	\includegraphics[width=\textwidth]{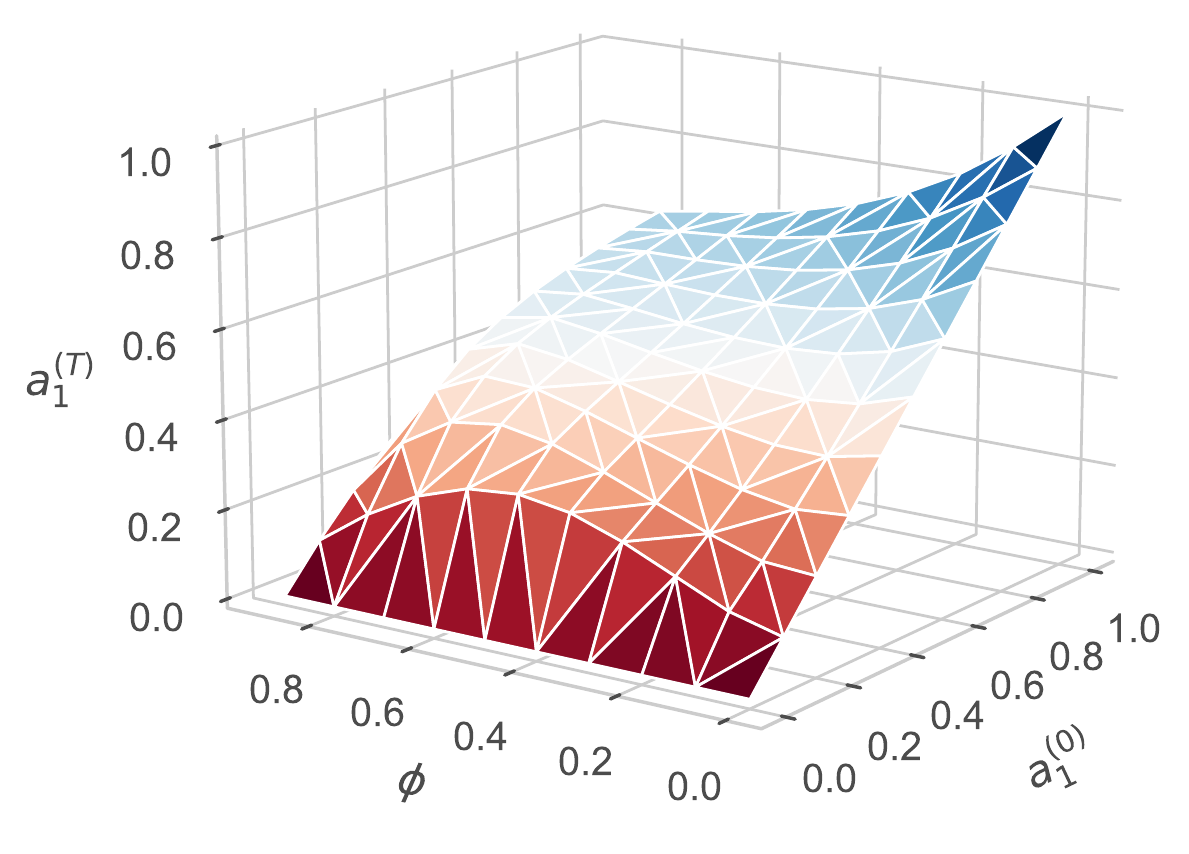}
	\caption{Re-scaled activation function}
	\label{fig:act_rescaled}
	\end{subfigure}
	
	\captionsetup{justification=justified}
	\caption{Final activation value (after performing $T=30$ iterations) of a single neuron using the quasi nonlinear reasoning rule with different transfer functions. In this simulation, we vary the nonlinearity degree and the raw activation value of the first neuron without activating the second one.}
\label{fig:act_rules}
\end{figure*}

Let us simulate to illustrate the behavior of the sigmoid function, the hyperbolic tangent function, and the new re-scaled transfer function. In this experiment, we couple these transfer functions with the quasi nonlinear reasoning rule introduced in Equation~\eqref{eq:reasoning-new}. Moreover, we use an FCM model comprised of two neurons with weights $w_{12}=0.8$, $w_{21}=0.5$, and zeros in the main diagonal. Figure~\ref{fig:act_rules} illustrates the activation values of the first neuron when activating it with random numbers in the $[0,1]$ interval and without activating the second one. The simulation results show that the sigmoid neuron is activated when its input is zero, while the hyperbolic tangent and the re-scaled transfer functions keep the neuron inactive. It can also be noticed that the surface generated with the re-scaled function is smoother than the hyperbolic tangent.

Another attractive property of the quasi nonlinear reasoning rule using the re-scaled transfer function is that activation values are not absolute but relative. In the model, the neurons' activation values depend not on the neurons' activation values in the previous iteration, the weights, and the initial conditions but the neurons' activation values in the current iteration. Overall, dividing each raw activation value by the Euclidean norm of the activation vector ensures that activation values are in the $[0,1]$ interval while promoting competition among the neurons. In the following section, we will mathematically study the convergence properties of the quasi nonlinear reasoning rule and the proposed re-scaled transfer function.

\section{Convergence analysis}
\label{sec:convergence:}

As mentioned in Section~\ref{sec:fcm}, FCM-based models can reach three possible states when it comes to their dynamic behavior: (i) fixed-point, (ii) limit cycle, or (iii) chaos. Likewise, these states apply to our model. In this section, we aim for convergence to diverse fixed-point attractors and to elude cycles, chaos, and unique fixed-point attractors.

A few FCM-based models have been mathematically studied in the literature with regards their convergence properties. The existence and uniqueness of fixed points of sigmoid FCMs was firstly discussed by Boutalis et al.~\cite{Boutalis2009} when the transfer function is given by $f(x) = 1/(1 + e^{-x})$ . Other theoretical results about this topic are the ones reported in Knight et al.~\cite{Knight2014}, and Harmati et al. \cite{Harmati2019}. As a generalization of the stability and instability properties of FCM-based models, in~\cite{Napoles2017} we introduced the definitions of $E$-stability and $E$-instability, and four sufficient conditions to fulfill these properties.

In~\cite{Leonardo2021}, the authors analyzed the behavior of FCMs from the perspective of their state spaces. More specifically, the research estimate bounds for the activation values of each neuron. Moreover, it was analyzed the coverage (i.e., the proportion of the induced activation space that is achievable by the neurons' activation values) and the proximity (i.e., the mean relative distance of the neurons' activation values at the borders of the feasible activation space) of neural concepts. The main theoretical result shows that the state space of any FCM equipped with an $F$-transfer function~\cite{Leonardo2021} shrinks infinitely with no guarantees of reaching a fixed point, but it does converge to the limit state space. Finally, in~\cite{Leonardo2020} it was proven that in an FCM-based model, a sigmoid neuron with self-feedback and no other incoming connection will always converge to a unique fixed-point regardless of its initial stimulus.

As a further addition to this research line, we will study the convergence features of the proposed reasoning mechanism. First, we establish the requirements for the appearance of cycles while examining some special situations. Then, we analyze the model conditions to avoid unique fixed-point attractors and the conditions to reach them.

\subsection{Basic notions and definitions}
\label{sec:convergence:basics}

Before presenting the main theoretical results devoted to the convergence of our model, let us re-write Equations \eqref{eq:reasoning-new} and \eqref{eq:re-scaled} using the following matrix-like notation:

\begin{equation}
\label{eq:FCMReasoningRule}
\mathbf{A}^{(t+1)} = \phi f \left(\mathbf{A}^{(t)}\mathbf{W} \right) + (1 - \phi)\mathbf{A}^{(0)}
\end{equation}

\noindent such that $f(.): \mathbb{R}^M \rightarrow \mathbb{R}^M $ is defined as follows:

\begin{equation}
\label{eq:transfer-function}
f(\mathbf{X}) =
\begin{cases}
\frac{\mathbf{X}}{\lvert\lvert \mathbf{X} \rvert\rvert_2} & \text{if $X \neq \overrightarrow{0}$}\\
0 & \text{otherwise.}
\end{cases}
\end{equation}


In our FCM-based model, we need to ensure that activation values lie in the $[0, 1]$ interval. Next, we will prove that $A^{(t+1)} \in [0, 1]^M$ by mathematical induction.

\begin{proof}
The fact that $\mathbf{A}^{(0)} \in [0, 1]^M$ can be considered a restriction of the proposed neural reasoning model. Let us assume that $\mathbf{A}^{(t)} \in [0, 1]^M$ and prove that $\mathbf{A}^{(t+1)} \in [0, 1]^M$.

Let us define that $\mathbf{A}^{(t)}\mathbf{W} = \mathbf{V}$ such that $\mathbf{V} = (v_1,v_2,\dots,v_M) \in \mathbb{R}^M$. Having said that $\mathbf{A}^{(t)} \in [0, 1]^M$ and that the elements of $\mathbf{W}$ are in $[0,1]$, the vector-matrix product produces non-negative components. Then, we will demonstrate that $f(\mathbf{V}) \in [0, 1]^M$ and that there are two cases. On the one hand, having $\mathbf{V} = \overrightarrow{0}$ implies that $f(\mathbf{V}) = \overrightarrow{0}$. On the other hand, having that $\mathbf{V} \neq \overrightarrow{0}$, we will demonstrate that $\frac{\mathbf{V}}{\lvert\lvert \mathbf{V} \rvert\rvert_2} \in [0, 1]^M$. Since we are using the Euclidean norm, we have that:

$$ \frac{\mathbf{V}}{\lvert\lvert \mathbf{V} \rvert\rvert_2} = \frac{(v_1,v_2,\dots,v_M)}{\sqrt{\sum_{i=1}^M v_i^2}} = \frac{(v_1,v_2,\dots,v_M)}{\sqrt{v_1^2 + v_2^2+ \dots + v_M^2}} $$

It can be verified that, for every $v_i$, the inequality

$$0 \leq \frac{v_i}{\sqrt{\sum_{i=1}^M v_i^2}} \leq 1$$

\noindent holds and $\frac{\mathbf{V}}{\lvert\lvert \mathbf{V} \rvert\rvert_2} \in [0, 1]^M$. Hence, we have that $\mathbf{A}^{(t+1)} = \phi f \left( \mathbf{V} \right) + $ $(1 - \phi)\mathbf{A}^{(0)}$, with $f(\mathbf{V}), \mathbf{A}^{(0)} \in [0, 1]^M$. Since $0 \leq \phi\leq 1$, the right-hand side yields a non-negative vector that reaches its maximum (component-wise) when $f(\mathbf{V}) = \mathbf{A}^{(0)} = (1, 1, \ldots, 1)$. Such a value for $\phi f \left( \mathbf{V} \right) + (1 - \phi)\mathbf{A}^{(0)}$ is equal to $\phi \left( 1, 1, \ldots, 1 \right) + (1 - \phi)(1, 1, \ldots, 1) = (1, 1, \ldots, 1)$. Finally, we obtain that $\mathbf{A}^{(t+1)} \in [0, 1]^M$ and this concludes our proof.

\end{proof} 

\subsection{Cyclic behavior analysis}
\label{sec:convergence:cycles}

As we mentioned earlier, the evasion of cycles is desired and we need to know when our system falls into them. Generally speaking, a cycle is reached when:
\\
\\
$\mathbf{A}^{(t+v)} = \mathbf{A}^{(t)}$ for any $v \in \mathbb{N}: v > 0$.
\\
\\
It should be noticed that when $v = 1$, we fall into a special kind of cycle: \textit{convergence}. If we expand this definition using the proposed reasoning rule, we have:

$$\phi f \left( \mathbf{A}^{(t+v-1)}\mathbf{W} \right) + (1 - \phi)\mathbf{A}^{(0)} = \phi f \left( \mathbf{A}^{(t-1)}\mathbf{W} \right) + (1 - \phi)\mathbf{A}^{(0)},$$

\noindent such that

$$\phi f \left( \mathbf{A}^{(t+v-1)}\mathbf{W} \right) = \phi f \left( \mathbf{A}^{(t-1)}\mathbf{W} \right).$$

At this point, we can put aside the case when $\phi = 0$, since it implies that $\mathbf{A}^{(t+1)} = \mathbf{A}^{(t)}$, $\forall t$, so convergence is attained. Simplifying on $\phi$ leads to the following:

\begin{equation}
\label{eq:cycles}
f \left( \mathbf{A}^{(t+v-1)}\mathbf{W} \right) = f \left( \mathbf{A}^{(t-1)}\mathbf{W} \right).
\end{equation}

Firstly, let us analyze the cases when $\mathbf{A}^{(t+v-1)}\mathbf{W} = \overrightarrow{0}$ and $\mathbf{A}^{(t-1)}\mathbf{W} = \overrightarrow{0}$. In general, it holds that, if $\mathbf{A}^{(t-1)}\mathbf{W} = \overrightarrow{0}$, then $\mathbf{A}^{(t)} = (1 - \phi) \mathbf{A}^{(0)}$ and $\mathbf{A}^{(t+1)} = \mathbf{A}^{(1)}$. Such a cycle is found by simplifying the following expression:

$$\mathbf{A}^{(t+1)} = \phi f \left( (1 - \phi) \mathbf{A}^{(0)} \mathbf{W} \right) + (1 - \phi)\mathbf{A}^{(0)}.$$

In contrast, if $\mathbf{A}^{(0)}W \neq \overrightarrow{0}$ and $\phi < 1$ (otherwise, the cycle would have been appeared earlier), then we have:

$$\mathbf{A}^{(t+1)} = \phi \frac{(1 - \phi) \mathbf{A}^{(0)} \mathbf{W}}{\lvert\lvert (1 - \phi) \mathbf{A}^{(0)} \mathbf{W} \rvert\rvert_2} + (1 - \phi)\mathbf{A}^{(0)}.$$

Since the norm of a vector multiplied by a scalar is equal to the scalar multiplied by the norm, we have that

$$\mathbf{A}^{(t+1)} = \phi \frac{(1 - \phi) \mathbf{A}^{(0)} W}{(1 - \phi) \lvert\lvert \mathbf{A}^{(0)} W \rvert\rvert_2} + (1 - \phi)\mathbf{A}^{(0)},$$

\noindent and

$$\mathbf{A}^{(t+1)} = \phi \frac{\mathbf{A}^{(0)} \mathbf{W}}{\lvert\lvert \mathbf{A}^{(0)} W \rvert\rvert_2} + (1 - \phi)\mathbf{A}^{(0)} = \mathbf{A}^{(1)}.$$

Next, we can use the nonzero branch of Equation \eqref{eq:transfer-function} to develop Equation \eqref{eq:cycles} as follows:

$$\frac{A^{(t+v-1)}W}{\lvert\lvert A^{(t+v-1)}W \rvert\rvert_2} = \frac{A^{(t-1)}W}{\lvert\lvert A^{(t-1)}W \rvert\rvert_2}.$$

Finally, we obtain that:

$$\left( \frac{\mathbf{A}^{(t+v-1)}}{\lvert\lvert \mathbf{A}^{(t+v-1)}\mathbf{W} \rvert\rvert_2} - \frac{\mathbf{A}^{(t-1)}}{\lvert\lvert \mathbf{A}^{(t-1)}\mathbf{W} \rvert\rvert_2} \right)\mathbf{W} = 0.$$

The satisfaction of this equality is difficult but still possible. This situation may occur: the vector resulting as the difference of the fractions is perpendicular to every column of $\mathbf{W}$ or such a difference is equal to $\overrightarrow{0}$. The former is less likely to happen, while the latter has a slightly higher probability of occurrence. This more probable case turns out when

$$\frac{\mathbf{A}^{(t+v-1)}}{\lvert\lvert \mathbf{A}^{(t+v-1)}\mathbf{W} \rvert\rvert_2} = \frac{\mathbf{A}^{(t-1)}}{\lvert\lvert \mathbf{A}^{(t-1)}\mathbf{W} \rvert\rvert_2},$$

\noindent and

$$\mathbf{A}^{(t+v-1)} = \frac{\lvert\lvert \mathbf{A}^{(t+v-1)}\mathbf{W} \rvert\rvert_2}{\lvert\lvert \mathbf{A}^{(t-1)}\mathbf{W} \rvert\rvert_2} \mathbf{A}^{(t-1)}.$$

The above means that $\mathbf{A}^{(t+v-1)}$ is equal to $\mathbf{A}^{(t-1)}$ multiplied by a scalar factor, which is the quotient of the Euclidean norms. Beyond the mathematical meaning, it implies that if an activation vector is multiple of another for the same initial stimulus, then the cyclic behavior appears.

\subsection{Unique fixed-points analysis}
\label{sec:convergence:unique-fixed-points}

When performing what-if simulations, it is desirable not to converge to unique fixed-point attractors. In these situations, reaching the same output regardless of the initial conditions must be avoided. Next, we analyze how the model parameters influence in the system stability properties. To do that, we split the task into two cases: $0 \leq \phi < 1$ and $\phi = 1$.
\\

\noindent \textbf{Case 1:} $0 \leq \phi < 1$. Our goal is to prove that there is no unique fixed-point attractor for all initial stimuli. Furthermore, in the process we found other interesting results.

Aiming at obtaining a more general result, let us define $g(.): \mathbb{R}^M \rightarrow \mathbb{R}^M$ as a continuous transfer function for our model. Also, the model's intrinsic restrictions for $\mathbf{W}$ (symmetry and weights in $[0, 1]$) and $\mathbf{A}^{(t)}$ (activation values in $[0, 1]$) are not required for our demonstration. This means that we will use real matrices and activation vectors whose only requirement is that the product $\mathbf{A}^{(t)}\mathbf{W}$ must be defined. Thus, the generic reasoning rule can be rewritten once again as follows:

\begin{equation}
\label{eq:GeneralReasoningRule}
\mathbf{A}^{(t+1)} = \phi g \left( \mathbf{A}^{(t)}\mathbf{W} \right) + (1 - \phi)\mathbf{A}^{(0)}
\end{equation}

\begin{theorem}[Injective convergence]
\label{the:injective-mapping}
In an FCM model using Equation \eqref{eq:GeneralReasoningRule}, when $0 \leq \phi < 1$, there are not two different initial stimuli leading to the same fixed-point attractor.
\end{theorem}


\begin{proof}
Let $\mathbf{A}_i^{(0)}, \mathbf{A}_j^{(0)} \in \mathbb{R}^M$ denote two unequal initial activation vectors and also assume that both lead to the same fixed-point attractor $\mathbf{V}$. Leaning on the mathematical analysis, we define that $\mathbf{V} = \lim_{t\to \infty}{\mathbf{A}_i^{(t)}} = \lim_{t\to \infty}{\mathbf{A}_j^{(t)}}$. 

The activation vectors $\mathbf{A}_i^{(0)}$ and $\mathbf{A}_j^{(0)}$ can be inserted in the Equation \eqref{eq:GeneralReasoningRule}, producing the following formulas:

$$\mathbf{A}_i^{(t+1)} = \phi g \left( \mathbf{A}_i^{(t)}\mathbf{W} \right) + (1 - \phi)\mathbf{A}_i^{(0)},$$

\noindent and

$$\mathbf{A}_j^{(t+1)} = \phi g \left( \mathbf{A}_j^{(t)}\mathbf{W} \right) + (1 - \phi)\mathbf{A}_j^{(0)}.$$

Also, after applying the limit when $t\to \infty$, we have that:

$$\lim_{t\to \infty}{\mathbf{A}_i^{(t+1)}} = \phi \lim_{t\to \infty} {g \left( \mathbf{A}_i^{(t)}W \right)} + (1 - \phi)\mathbf{A}_i^{(0)},$$
\noindent and
$$\lim_{t\to \infty}{\mathbf{A}_j^{(t+1)}} = \phi \lim_{t\to \infty} {g \left( \mathbf{A}_j^{(t)}\mathbf{W} \right)} + (1 - \phi)\mathbf{A}_j^{(0)}.$$

Given the continuity of $g(.)$ and using the basic properties of limit operations, it yields:

$$\lim_{t\to \infty}{\mathbf{A}_i^{(t+1)}} = \phi g \left( \lim_{t\to \infty}{\mathbf{A}_i^{(t)}}\mathbf{W} \right) + (1 - \phi)\mathbf{A}_i^{(0)},$$
\noindent and
$$\lim_{t\to \infty}{\mathbf{A}_j^{(t+1)}} = \phi g \left( \lim_{t\to \infty}{\mathbf{A}_j^{(t)}}\mathbf{W} \right) + (1 - \phi)\mathbf{A}_j^{(0)}.$$

Now, substituting all limits by its value and solving on $\mathbf{A}_i^{(0)}$ and $\mathbf{A}_j^{(0)}$ respectively, we obtain:

$$\mathbf{A}_i^{(0)} = \frac{\mathbf{V} - \phi g (\mathbf{V} \mathbf{W})}{1 - \phi},$$
\noindent and
$$\mathbf{A}_j^{(0)} = \frac{\mathbf{V} - \phi g (\mathbf{V} \mathbf{W})}{1 - \phi}.$$

Note that the formulas above are not undefined since $\phi \neq 1$. At this point, the assumption $\mathbf{A}_i^{(0)} \neq \mathbf{A}_j^{(0)}$ leads us to $\mathbf{A}_i^{(0)} = \mathbf{A}_j^{(0)}$. Therefore, we reach a contradiction and the proof by \textit{reductio ad absurdum} is completed.
\end{proof}

Theorem \ref{the:injective-mapping} brings a straightforward consequence, which is portrayed in the following corollary.

\begin{corollary}[Relaxed injective convergence]
\label{cor:no-unique}
In an FCM model using Equation \eqref{eq:GeneralReasoningRule}, when $0 \leq \phi < 1$, there is no unique fixed-point attractor for all initial stimuli.
\end{corollary}

Theorem \ref{the:injective-mapping} and Corollary \ref{cor:no-unique} are valid when using the Equation \eqref{eq:GeneralReasoningRule}, but such a general equation differs from Equation \eqref{eq:FCMReasoningRule}. Hence, we need to analyze the continuity of $f(.)$. 

On the one hand, the first branch concerning $\frac{\mathbf{X}}{\lvert\lvert \mathbf{X} \rvert\rvert_2}$ is continuous since it is a elementary function that never gets undefined when $\mathbf{X} \neq \overrightarrow{0}$. On the other hand, $f(\overrightarrow{0}) = \overrightarrow{0}$, and for this branch the continuity also requires that:

$$\lim_{\mathbf{X} \to \overrightarrow{0}}{f(\mathbf{X})} = \overrightarrow{0}.$$

Next, we will prove that the previous expression does not hold by using the limit on $\mathcal{D} = \{(u,u,\dots,u) \in \mathbb{R}^M : u \neq 0 \}$, such that ${\mathcal{D}} \subset Dom  f(.)$ and $Dom f(.)$ is the domain of $f(.)$. Then, if $\mathbf{X} \in {\cal D}$ we have that the following:

$$f(X) = \frac{(u,u,\dots,u)}{\sqrt{u^2 + u^2+ \dots + u^2}} = \frac{(u,u,\dots,u)}{u\sqrt{M}}.$$

\noindent Now, the limit on ${\cal D}$ is calculated as follows:
$$\lim_{\mathbf{X} \to \overrightarrow{0}}{f(\mathbf{X})} = \lim_{u \to 0}{\frac{(u,u,\dots,u)}{u\sqrt{M}}} = (\frac{1}{\sqrt{M}}, \frac{1}{\sqrt{M}}, \dots, \frac{1}{\sqrt{M}}) \neq \overrightarrow{0},$$

\noindent which implies that $f(.)$ is not continuous at $\overrightarrow{0}$.

As we know from the cyclic behavior analysis in Section \ref{sec:convergence:cycles}, if after some initial stimulus we get $\mathbf{A}^{(t)}\mathbf{W} = \overrightarrow{0}$ for some $t$, then our model will exhibit cyclic behavior. Similarly, if $f(.)$ is evaluated on the discontinuity point $\overrightarrow{0}$ during the inference process, a cycle will be reached. Since the Theorem \ref{the:injective-mapping} and the Corollary \ref{cor:no-unique} conclude on the nonexistence of stimuli (more than one) leading to a fixed-point attractor, then the cyclic behavior is consistent with such a conclusion.

Finally, the theorem and the corollary are also valid using the Equation \eqref{eq:FCMReasoningRule} instead of Equation \eqref{eq:GeneralReasoningRule}.
The following theorem and corollary summarize the results exposed.

\begin{theorem}
\label{the:injective-mapping-eq1}
In an FCM model using the Equation \eqref{eq:FCMReasoningRule}, when $0 \leq \phi < 1$ there is no couple of different initial stimuli leading to the same fixed-point attractor.
\end{theorem}

\begin{corollary}
\label{cor:no-unique-eq1}
In the model described by the reasoning rule in the Equation \eqref{eq:FCMReasoningRule}, when $0 \leq \phi < 1$ there is no unique fixed-point attractor for all initial stimuli.
\end{corollary}

Summarizing, the obtained results for the case $0 \leq \phi < 1$ allow performing what-if simulations without the concerns of converging to a unique fixed point.

\noindent \textbf{Case 2:} $\phi = 1$. We establish the equivalence between our reasoning rule and the \textit{power iteration method} \cite{MisesPraktischeVD} by using matrix properties. This leads us to interesting conclusions about the convergence of our model. For the case in point, the reasoning rule is reduced to the following expression:

\begin{equation}
\mathbf{A}^{(t+1)} = f \left( \mathbf{A}^{(t)}\mathbf{W} \right).
\end{equation}

We could exclude the case when $\mathbf{A}^{(t)} \mathbf{W} = \overrightarrow{0}$ for some $t$, for it causes the model to converge to $\overrightarrow{0}$. Therefore, we rewrite the previous formula as follows:

\begin{equation}
\label{eq:powerIterationReasoning}
\mathbf{A}^{(t+1)} = \frac{\mathbf{A}^{(t)}\mathbf{W}}{\vert\vert \mathbf{A}^{(t)}\mathbf{W} \vert\vert_2}.
\end{equation}

Note that the dimensions of $\mathbf{A}^{(t)}$ and $\mathbf{W}$ are $1$x$M$ and $M$x$M$, respectively. By transposing the elements in Equation \eqref{eq:powerIterationReasoning}, we change the order of the vectors, not the results. After applying the property of the transpose of a matrix product and the fact that the Euclidean norm of a matrix does not change after transposition, we obtain the following:

\begin{equation}
{\mathbf{A}^{(t+1)}}^\top = \frac{\mathbf{W}^\top{A^{(t)}}^\top}{\vert\vert \mathbf{W}^\top{\mathbf{A}^{(t)}}^\top \vert\vert_2}.
\end{equation}

\noindent Given that the weight matrix $W$ derived from the correlation patterns is symmetric, it yields that:

\begin{equation}
\label{power-iteration}
{\mathbf{A}^{(t+1)}}^\top = \frac{\mathbf{W}{\mathbf{A}^{(t)}}^\top}{\vert\vert \mathbf{W}{\mathbf{A}^{(t)}}^\top \vert\vert_2}.
\end{equation}

Equation \eqref{power-iteration} matches with the power iteration method \cite{MisesPraktischeVD}. Given a diagonalizable matrix $\mathbf{W}$, the algorithm will produce a number $\lambda$, which is the greatest (in absolute value) eigenvalue of $\mathbf{W}$, and a nonzero vector $\mathbf{v}$, which is the corresponding eigenvector of $\lambda$, that is, $\mathbf{W} \mathbf{v} = \lambda \mathbf{v}$. The algorithm is also known as the power method or the Von Mises iteration \cite{MisesPraktischeVD} and starts with a vector $\mathbf{b}_0$, which may be an approximation to the dominant eigenvector or just a random vector. In every iteration, the vector $\mathbf{b}_t$ is multiplied by $\mathbf{W}$ and normalized. The algorithm is described by the recurrence relation:

\begin{equation}
\mathbf{b}_{t+1} = \frac{\mathbf{W} \mathbf{b}_t}{\vert \vert \mathbf{W} \mathbf{b}_t \vert \vert}.
\end{equation}

If we assume that $W$ has an eigenvalue that is strictly greater in magnitude than its other eigenvalues and that the starting vector $\mathbf{b}_0$ has a nonzero component in the direction of an eigenvector associated with the dominant eigenvalue, then the subsequence ($\mathbf{b}_t$) converges to an eigenvector associated with the dominant eigenvalue. Such an eigenvector always exists and it is unique provided that these two assumptions are fulfilled; otherwise, the sequence might not converge.

In order to match the terminology used in the recurrence relation of the power iteration with our terms, we establish that $\mathbf{b}_t = {\mathbf{A}^{(t)}}^\top$. Moreover, the symmetry of $W$ implies that it is diagonalizable and there are $M$ real eigenvalues. Summarizing, the case $\phi = 1$ allows modeling invariant situations in which the model needs to converge to the same fixed-point attractor regardless of the initial conditions.


\section{Numerical simulations}
\label{sec:simulations}

In this section, we conduct numerical simulations using three scenarios involving protected features reported in the literature. Also, we provide experimental evidence further supporting the main theoretical results of this research.

\subsection{German credit dataset}
\label{sec:simulations:dataset}

Aiming at testing our FCM-based model, we use the German Credit dataset~\cite{Dua:2019} that is widely used in the context of fairness in machine learning and classifies loan applicants as good or bad credit risks. Features are treated either as numeric or nominal following the data description at the UCI Machine Learning Repository \cite{Dua:2019}. Ordinal variables are treated as nominal as it is not clear how to order the categories within some variables. Treatment of ordinal variables should be decided by domain experts who are familiar with the data-generating procedure and the decision making rationale. It is worth mentioning that features \textit{age}, \textit{foreign worker} and \textit{gender} are considered to be protected following similar practices in the literature \textit{gender}, \textit{foreign worker} and \textit{age}~\cite{aif3602018,bonchi2017exposing,delobelle2021ethical, verma2018fairness, zemel2013learning, friedler2019comparative}.

During pre-processing steps, (i) the nominal features \emph{sex \& marital status} are re-coded to include information only related to gender following~\cite{aif3602018}'s analysis and (ii) numeric features are normalized in the $[0,1]$ interval. The FCM's weight matrix is computed as described in Section \ref{sec:proposal:model:correlation} where each entry represents the absolute correlation between every pair of variables.

Table \ref{tab:dataset} shows the features associated with the German Credit dataset, the absolute correlation values between the problem features and each protected feature (highlighted in gray) and the FRU values as reported in ~\cite{Napoles2021Afuzzyrough}, which provide insight into the explicit bias patterns (i.e., how the boundary regions change when suppressing a given feature). The star symbol is used to denote significant correlation results.

\begin{table}[!ht]
\caption{Correlation\textsuperscript{1} values between protected and unprotected features, and FRU values that provide information about the explicit bias.}
\resizebox{0.988\linewidth}{!}{
\begin{tabular}{l|l|ccc|c}
\hline
\multicolumn{1}{c|}{\multirow{2}{*}{ID}} & \multicolumn{1}{c|}{\multirow{2}{*}{Features}} & \multicolumn{3}{c|}{Correlation with}  & \multirow{2}{*}{FRU} \\ \cline{3-5}
\multicolumn{1}{c|}{}  & \multicolumn{1}{c|}{}  & \multicolumn{1}{c|}{Age}   & \multicolumn{1}{c|}{Foreign worker} & \multicolumn{1}{c|}{Gender} &  \\ \hline
\rowcolor{Gray} f1 & Age & \multicolumn{1}{c|}{1.0*}  & \multicolumn{1}{c|}{0} & 0.03*  & 0.11 \\
f2 & Credit amount & \multicolumn{1}{c|}{0.03} & \multicolumn{1}{c|}{0} & 0.01*  & 0.07 \\
f3 & Credit history & \multicolumn{1}{c|}{0.03*} & \multicolumn{1}{c|}{0.07} & 0.12*  & 0.26 \\
f4 & Months & \multicolumn{1}{c|}{0.04} & \multicolumn{1}{c|}{0.02*} & 0.01 & 0.11 \\
\rowcolor{Gray} f5 & Foreign worker & \multicolumn{1}{c|}{0.0} & \multicolumn{1}{c|}{1.0*} & 0.04 & 0.09 \\
f6 & Housing & \multicolumn{1}{c|}{0.09*} & \multicolumn{1}{c|}{0.07} & 0.23* & 0.17 \\
f7 & Installment rate & \multicolumn{1}{c|}{0.06} & \multicolumn{1}{c|}{0.01*} & 0.01 & 0.2 \\
f8 & Job & \multicolumn{1}{c|}{0.03*} & \multicolumn{1}{c|}{0.1*} & 0.09* & 0.23 \\
f9 & Existing credits & \multicolumn{1}{c|}{0.15*} & \multicolumn{1}{c|}{0.0} & 0.01*  & 0.23 \\
f10 & People liable & \multicolumn{1}{c|}{0.01*} & \multicolumn{1}{c|}{0.07*} & 0.2* & 0.14 \\
f11 & Other debtors & \multicolumn{1}{c|}{0.0} & \multicolumn{1}{c|}{0.12*} & 0.01 & 0.15 \\
f12 & Other installment & \multicolumn{1}{c|}{0.0} & \multicolumn{1}{c|}{0.04} & 0.05 & 0.23 \\
\rowcolor{Gray} f13 & Gender & \multicolumn{1}{c|}{0.03*} & \multicolumn{1}{c|}{0.04} & 1.0* & 0.22 \\
f14 & Employment since & \multicolumn{1}{c|}{0.17*} & \multicolumn{1}{c|}{0.08} & 0.22* & 0.36 \\
f15 & Residence since & \multicolumn{1}{c|}{0.27*} & \multicolumn{1}{c|}{0.0} & 0.0 & 0.19 \\
f16 & Property & \multicolumn{1}{c|}{0.05*} & \multicolumn{1}{c|}{0.14*} & 0.09* & 0.34 \\
f17 & Purpose & \multicolumn{1}{c|}{0.03*} & \multicolumn{1}{c|}{0.17*} & 0.15* & 0.37 \\
f18 & Savings account & \multicolumn{1}{c|}{0.01}  & \multicolumn{1}{c|}{0.04} & 0.07 & 0.32 \\
f19 & Checking account & \multicolumn{1}{c|}{0.01}  & \multicolumn{1}{c|}{0.08} & 0.03   & 0.39 \\
f20 & Telephone & \multicolumn{1}{c|}{0.02*} & \multicolumn{1}{c|}{0.1*} & 0.07* & 0.22 \\ \hline
\end{tabular}}
\label{tab:dataset}
\scriptsize\textsuperscript{1} Pearson correlation is used between numeric features and Cram\'er's V is used between nominal features such that the star indicates significant p-value ($p < 0.05$). The R-squared coefficient of determination is used between numeric and nominal features with the star indicating a significant F-statistic (larger than the critical value).
\end{table}

\begin{figure}[!htbp]
    \centering
    \resizebox{0.9\linewidth}{!}{
    \includegraphics{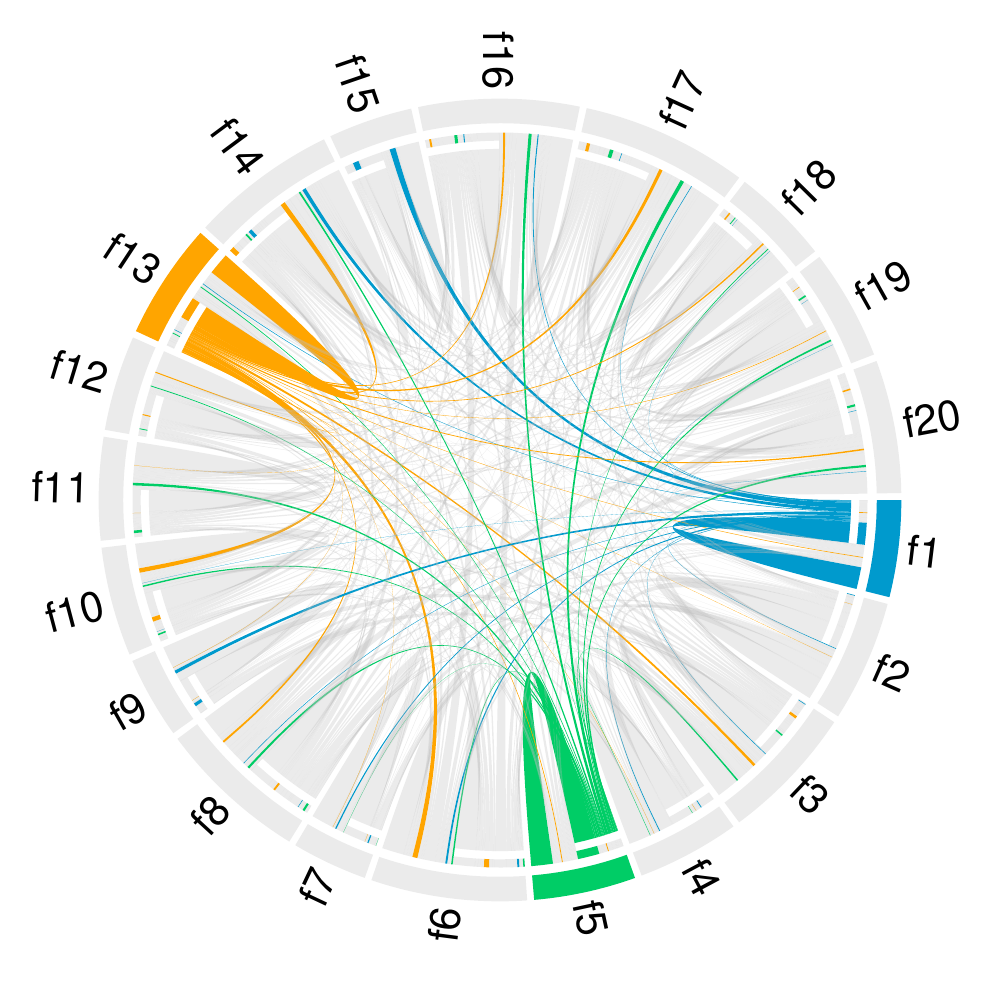}}
    \caption{Pairwise correlations between features in the German Credit dataset. The colored chords show the direct interactions between protected and unprotected features, while gray chords denote the interaction among unprotected features. Overall, this graph shows the complex interactions among the features when modeling implicit bias.}
    \label{fig:chord}
\end{figure}

Figure~\ref{fig:chord} provides an overview of the correlation patterns in an attempt to visualize all possible pathways through which implicit bias related to the sensitive features can be propagated within the system. In this chord chart, we highlighted the correlation between each unprotected feature and the protected ones. However, our FCM model takes into account all interactions when quantifying implicit bias.

\subsection{Results and discussion}
\label{sec:simulations:discussion}

Leveraging the simulation capability of the proposed FCM-based model, we perform a what-if analysis in three different scenarios. Moreover, we will illustrate how the nonlinear parameter in Equation \eqref{eq:reasoning-new} impacts the results and discuss how it can be used as a powerful tool to perform richer what-if simulations that can be useful for decision makers. 

As mentioned, these what-if simulations can be performed by (i) assigning initial values to selected concepts denoting unprotected features, (ii) executing the inference mechanism of the FCM model, and (iii) observing the resulting activation values of concepts denoting protected features as a way to quantify implicit bias. The scenarios to be investigated and the structure of initial activation vectors are detailed below. We generate 20 activation vectors for each scenario where concepts denoting selected unprotected features are randomly initialized, whereas the remaining ones are set to zero.

\begin{itemize}
    \item {\textbf{Scenario 1}. We activate three neural concepts representing unprotected features having large absolute correlation with the protected feature \textit{age}. These unprotected features are \textit{existing credits} (f9), \textit{employment since} (f14) and \textit{residence since} (f15), which encode the set of initial activation vectors of the form: $A^{(0)}=(0, 0, 0, 0, 0, 0, 0, 0,$ $f9, 0, 0, 0, 0, f14, f15, 0, 0, 0, 0, 0)$ where f9, f14 and f15 are uniform random values in $(0,1]$.} 
    \item {\textbf{Scenario 2}. We activate three neural concepts representing unprotected features having large absolute correlation with the protected feature \textit{foreign worker}. These unprotected features are \textit{other debtors} (f11), \textit{property} (f16) and \textit{purpose} (f17), which encode the set of initial activation vectors of the form: $A_1^{(0)}=(0, 0, 0, 0, 0, 0, 0, 0, 0, 0,$ $f11, 0, 0, 0, 0, f16, f17, 0, 0, 0)$ where f11, f16 and f17 are uniform random values in $(0,1]$.} 
    \item {\textbf{Scenario 3}. We activate four concepts representing unprotected features having large absolute correlation with the protected feature \textit{gender}. These unprotected features are \textit{housing} (f6), \textit{people liable} (f10), \textit{employment since} (f14) and \textit{purpose} (f17), which encode the set of initial activation vectors of the form: $A_1^{(0)}=(0, 0, 0, 0, 0,f6, 0, 0, 0, $ $f10, 0, 0, 0, f14, 0, 0, f17, 0, 0, 0)$ where f6, f10, f14 and f17 are uniform random values in $(0,1]$.} 
\end{itemize}

\begin{figure*}[!htbp]
\center

    \begin{subfigure}{0.33\textwidth}
	\center
	\includegraphics[width=\textwidth]{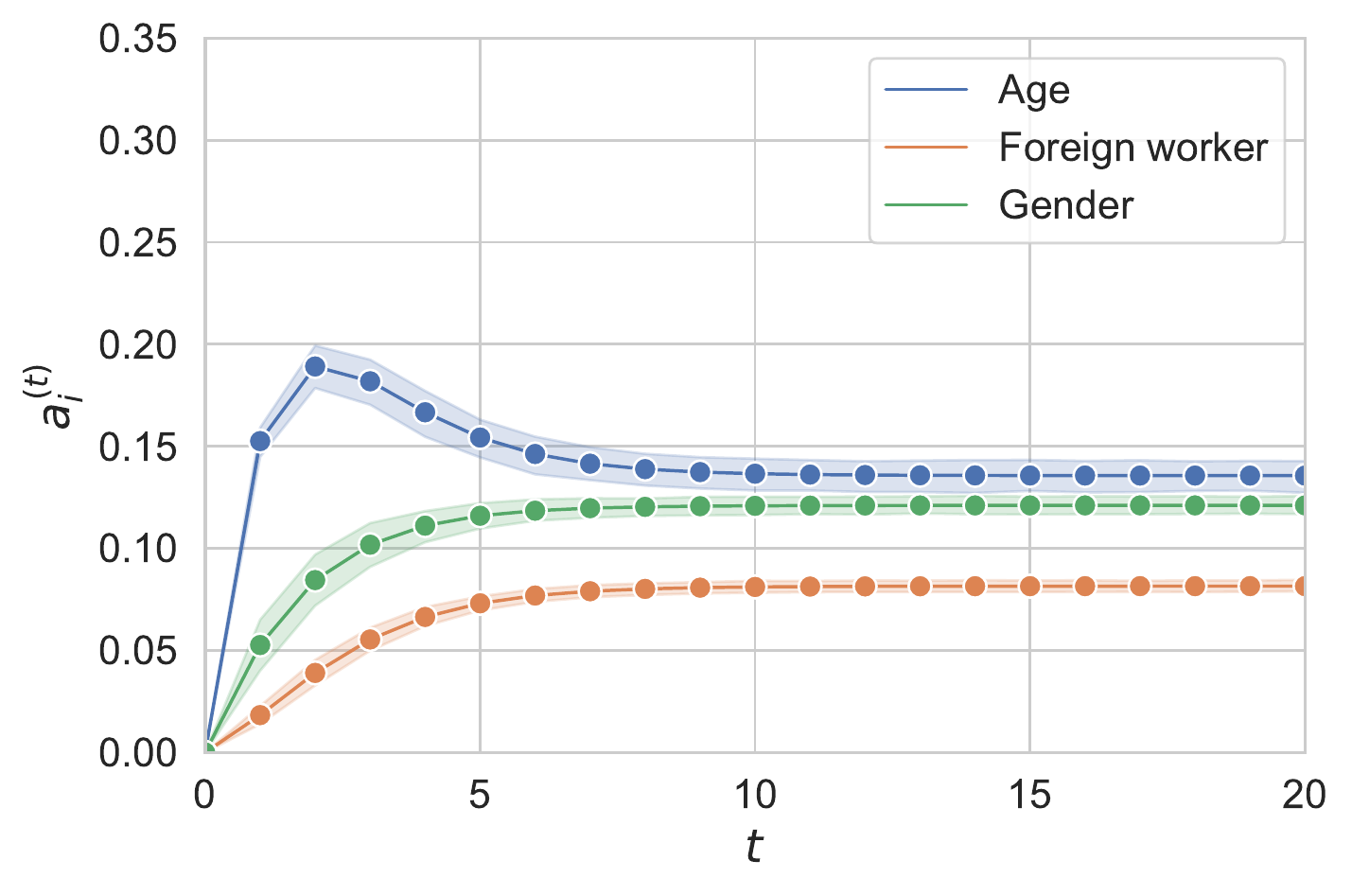}
	\caption{$\phi=0.6$}
	\end{subfigure}
	\begin{subfigure}{0.33\textwidth}
	\center
	\includegraphics[width=\textwidth]{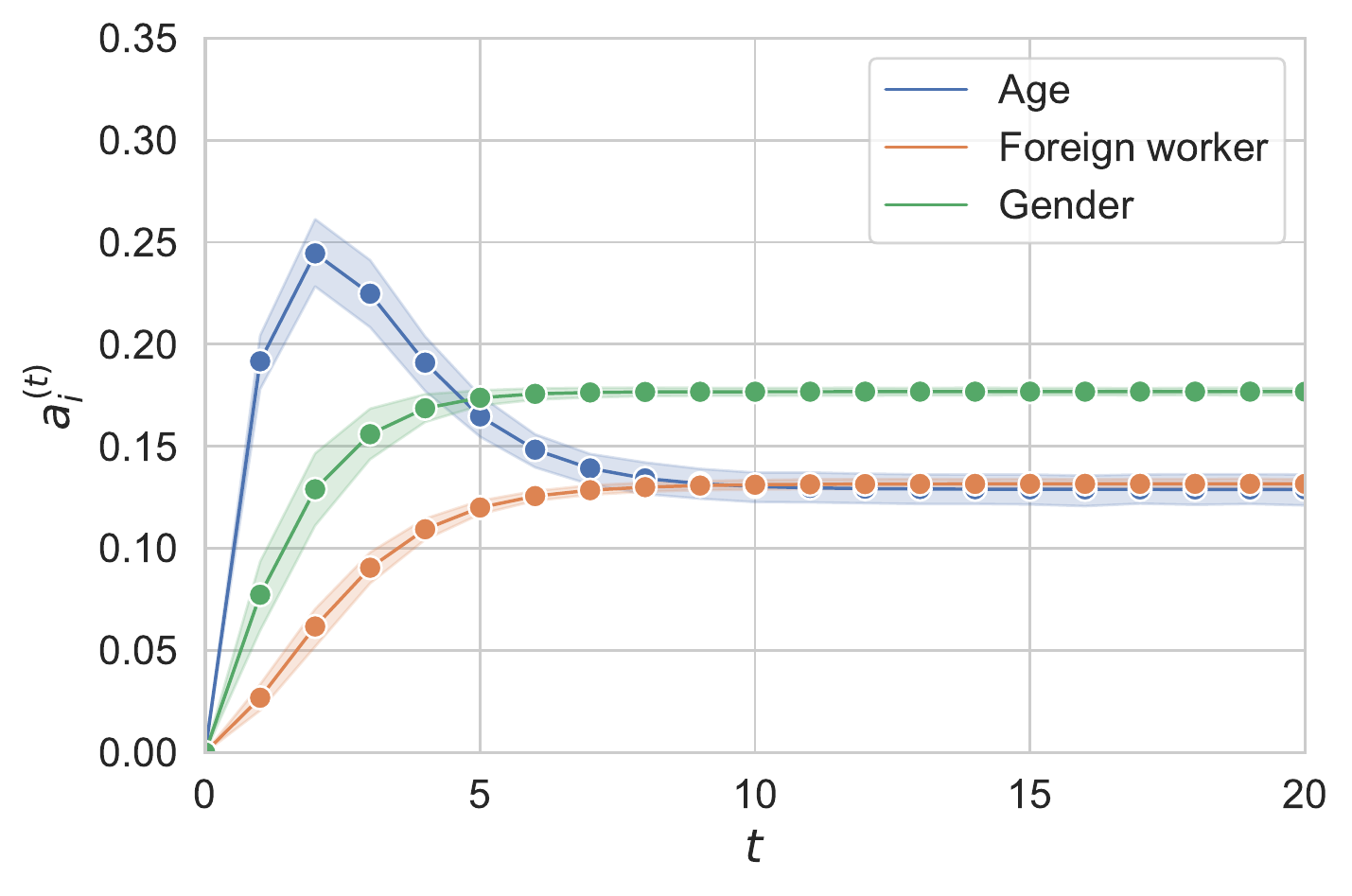}
	\caption{$\phi=0.8$}
	\end{subfigure}
	\begin{subfigure}{0.33\textwidth}
	\center
	\includegraphics[width=\textwidth]{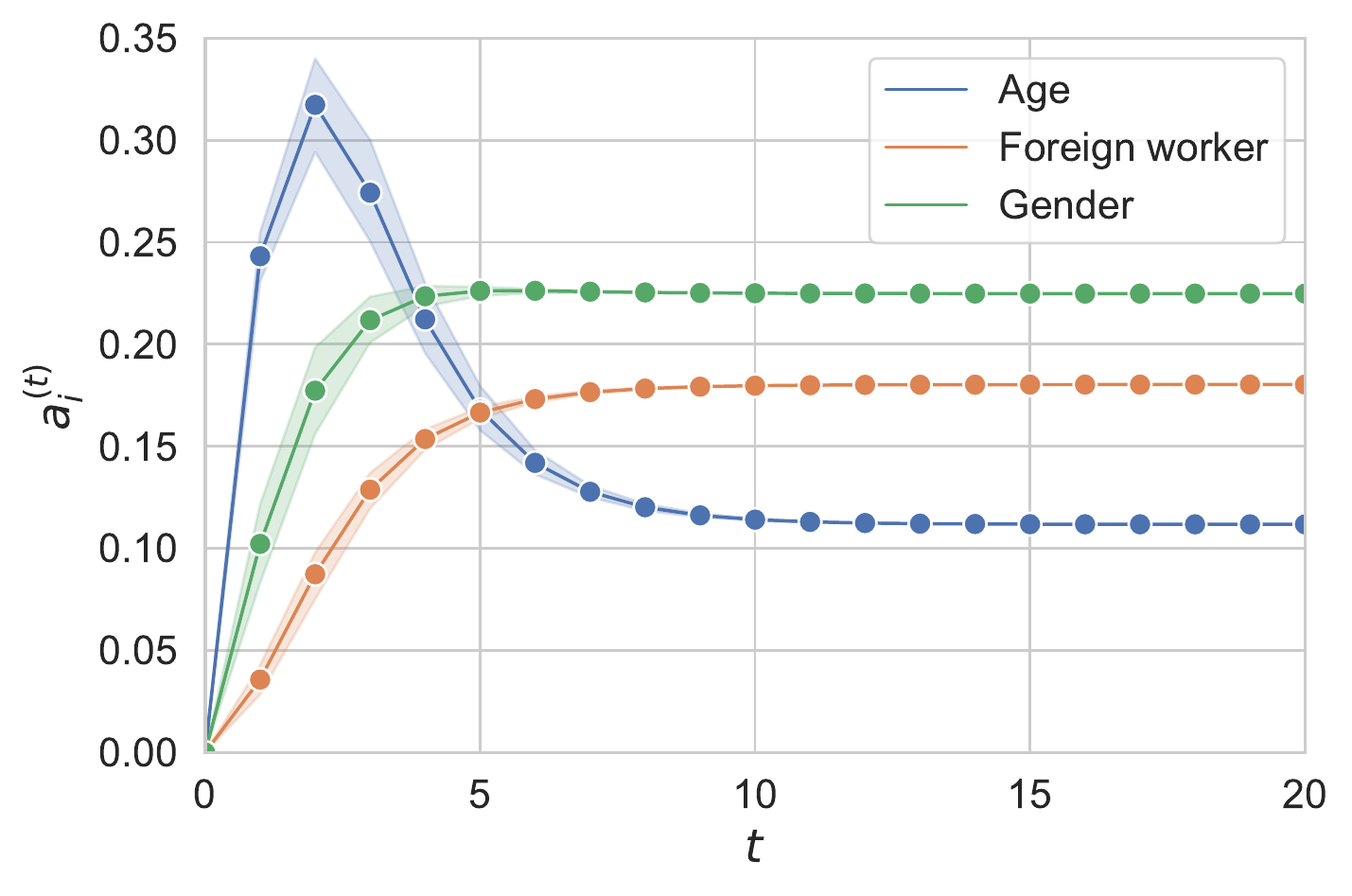}
	\caption{$\phi=1.0$}
	\end{subfigure}
	
	\captionsetup{justification=justified}
	\caption{Activation values of neurons denoting protected features for the first scenario. The unprotected features \textit{existing credits}, \textit{employment since} and \textit{residence since}, which have the largest correlation with the protected feature \textit{age}, are randomly initialized when running the simulations.}
	
\label{fig:phi_var_age}
\end{figure*}

\begin{figure*}[!htbp]
\center

    \begin{subfigure}{0.33\textwidth}
	\center
	\includegraphics[width=\textwidth]{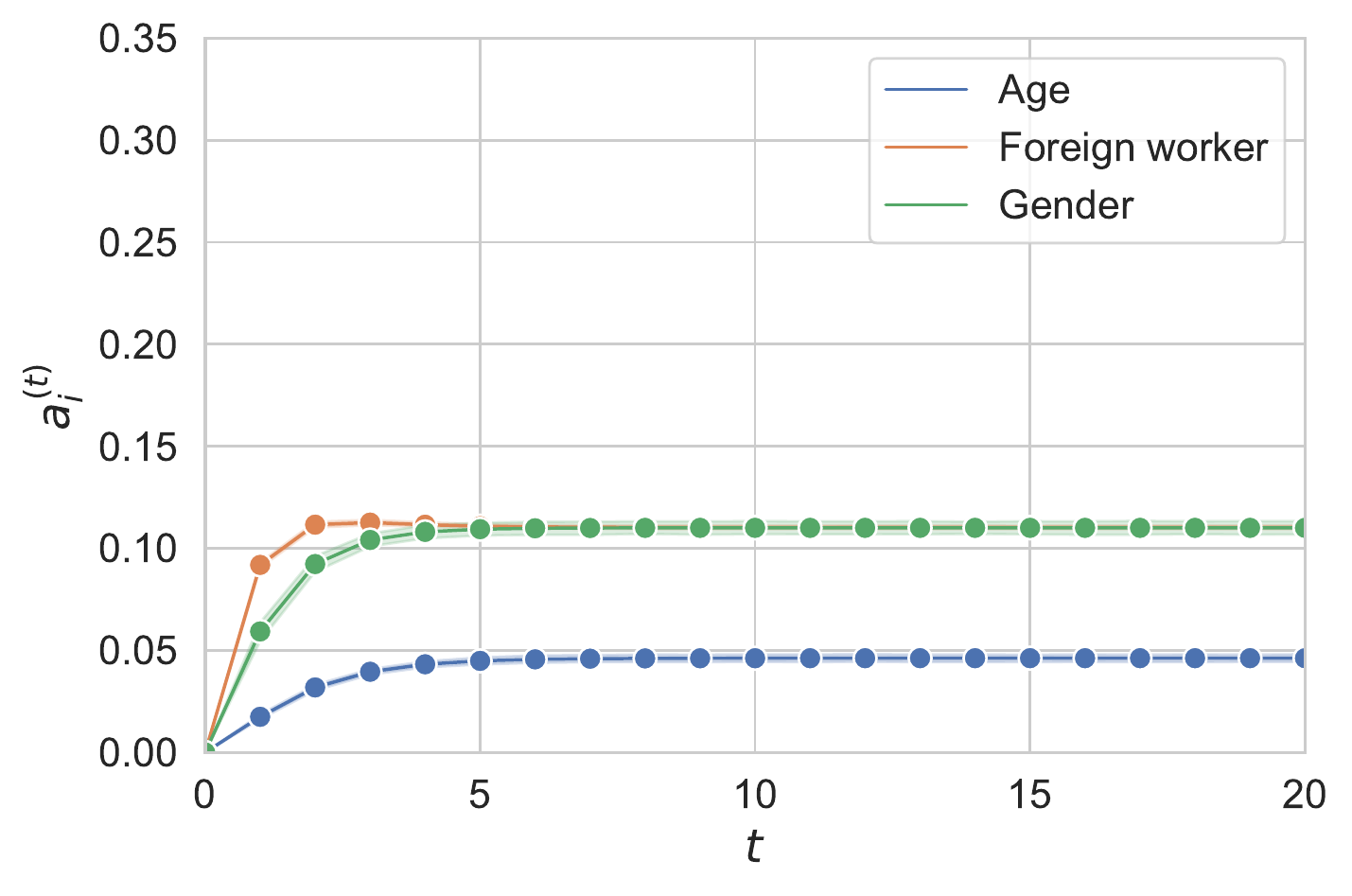}
	\caption{$\phi=0.6$}
	\end{subfigure}
	\begin{subfigure}{0.33\textwidth}
	\center
	\includegraphics[width=\textwidth]{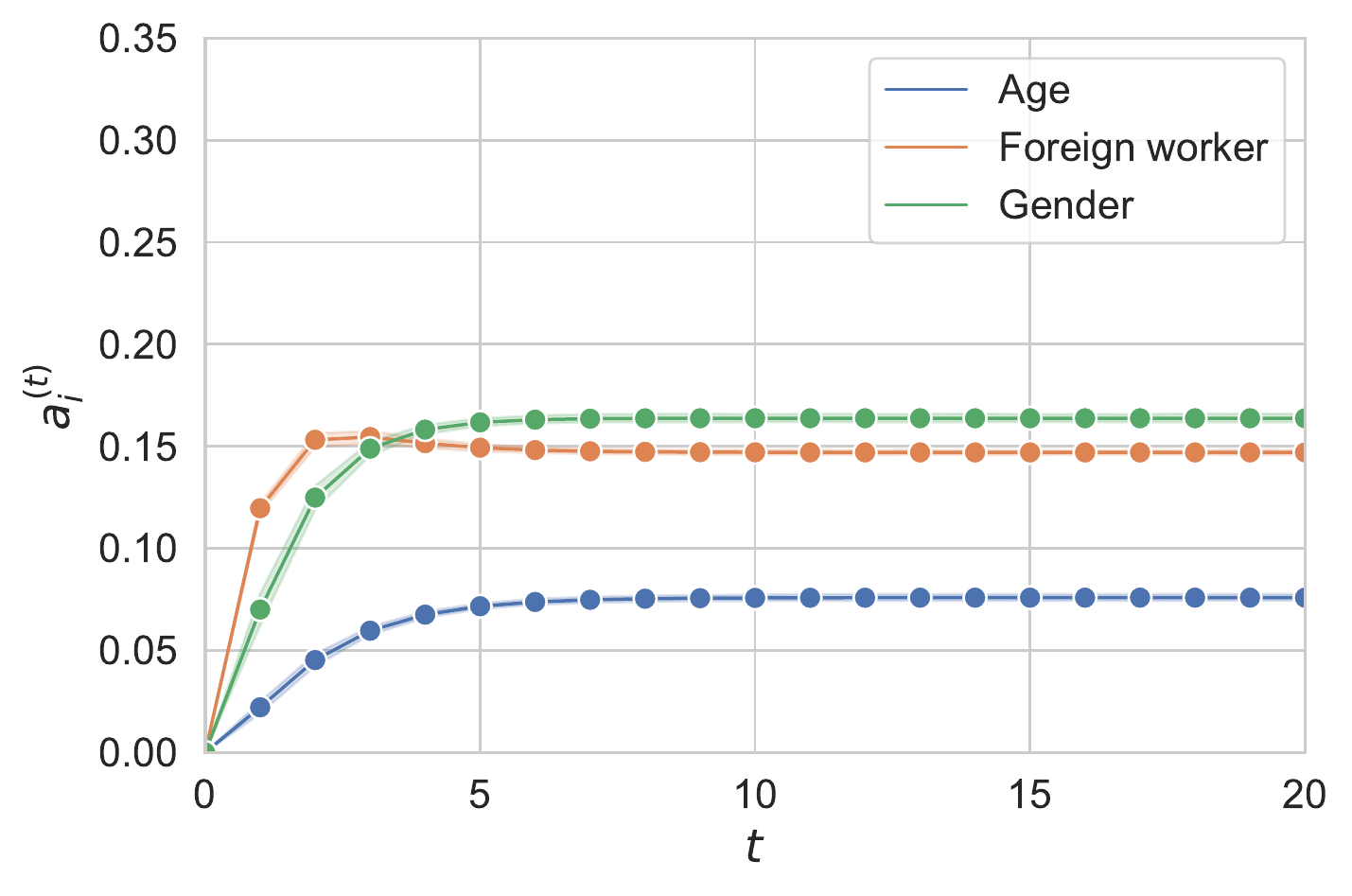}
	\caption{$\phi=0.8$}
	\end{subfigure}
	\begin{subfigure}{0.33\textwidth}
	\center
	\includegraphics[width=\textwidth]{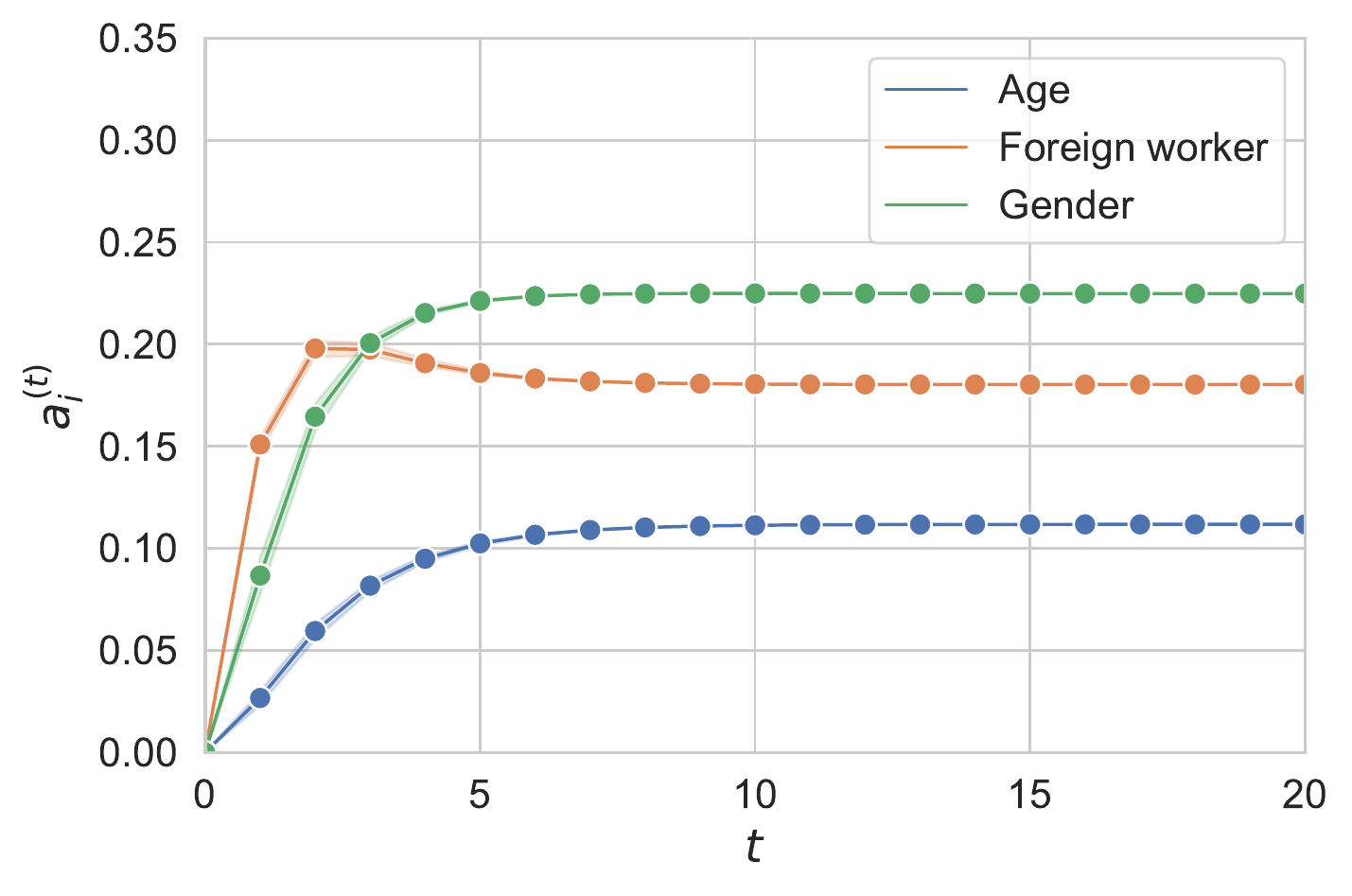}
	\caption{$\phi=1.0$}
	\end{subfigure}
	
	\captionsetup{justification=justified}
	\caption{Activation values of neurons denoting protected features for the second scenario. The unprotected features \textit{other debtors}, \textit{property} and \textit{purpose}, which have the largest correlation with the protected feature \textit{foreign worker}, are randomly initialized when running the simulations.}
	
\label{fig:phi_var_foreign}
\end{figure*}

\begin{figure*}[!htbp]
\center

    \begin{subfigure}{0.33\textwidth}
	\center
	\includegraphics[width=\textwidth]{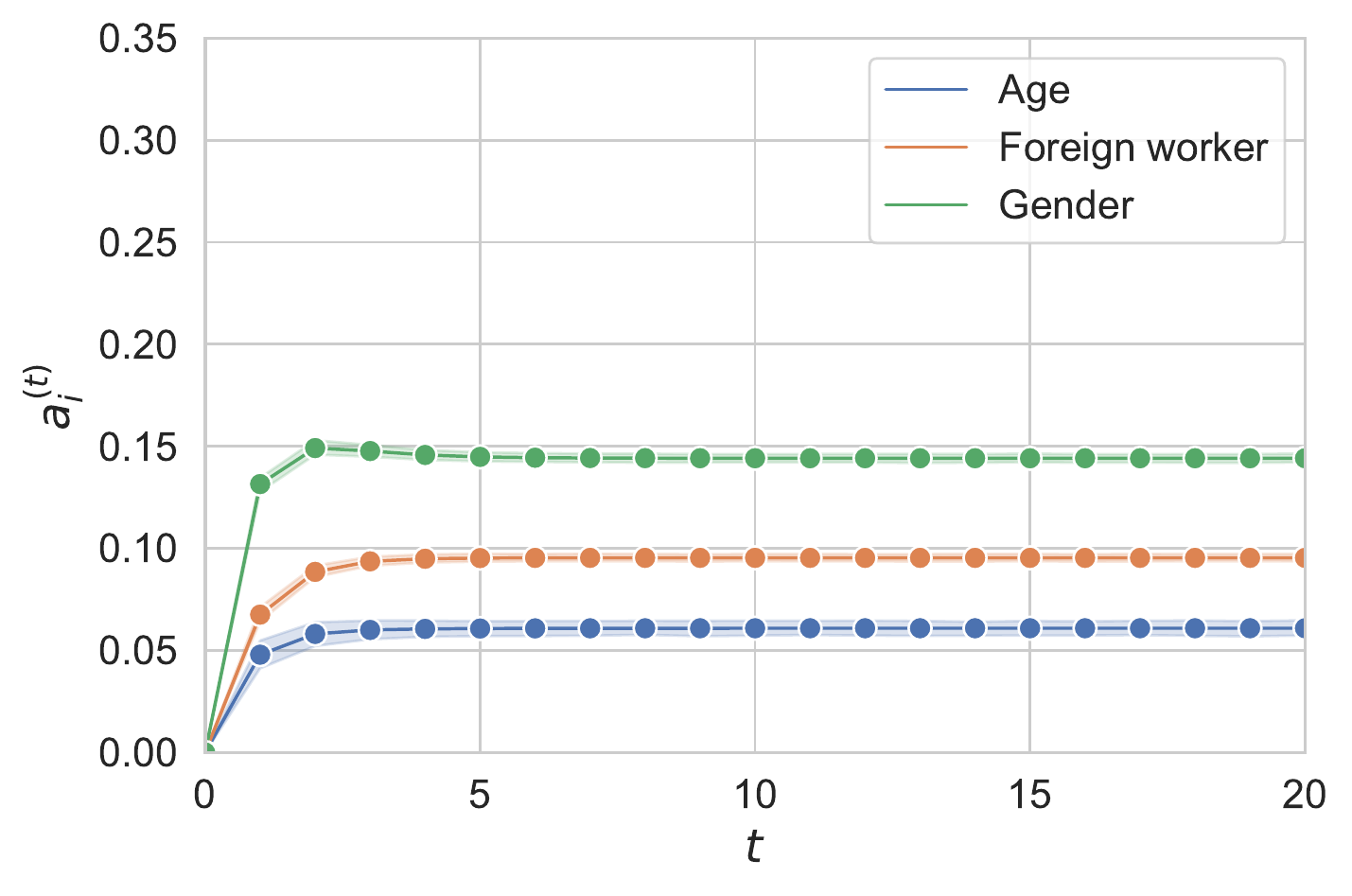}
	\caption{$\phi=0.6$}
	\end{subfigure}
	\begin{subfigure}{0.33\textwidth}
	\center
	\includegraphics[width=\textwidth]{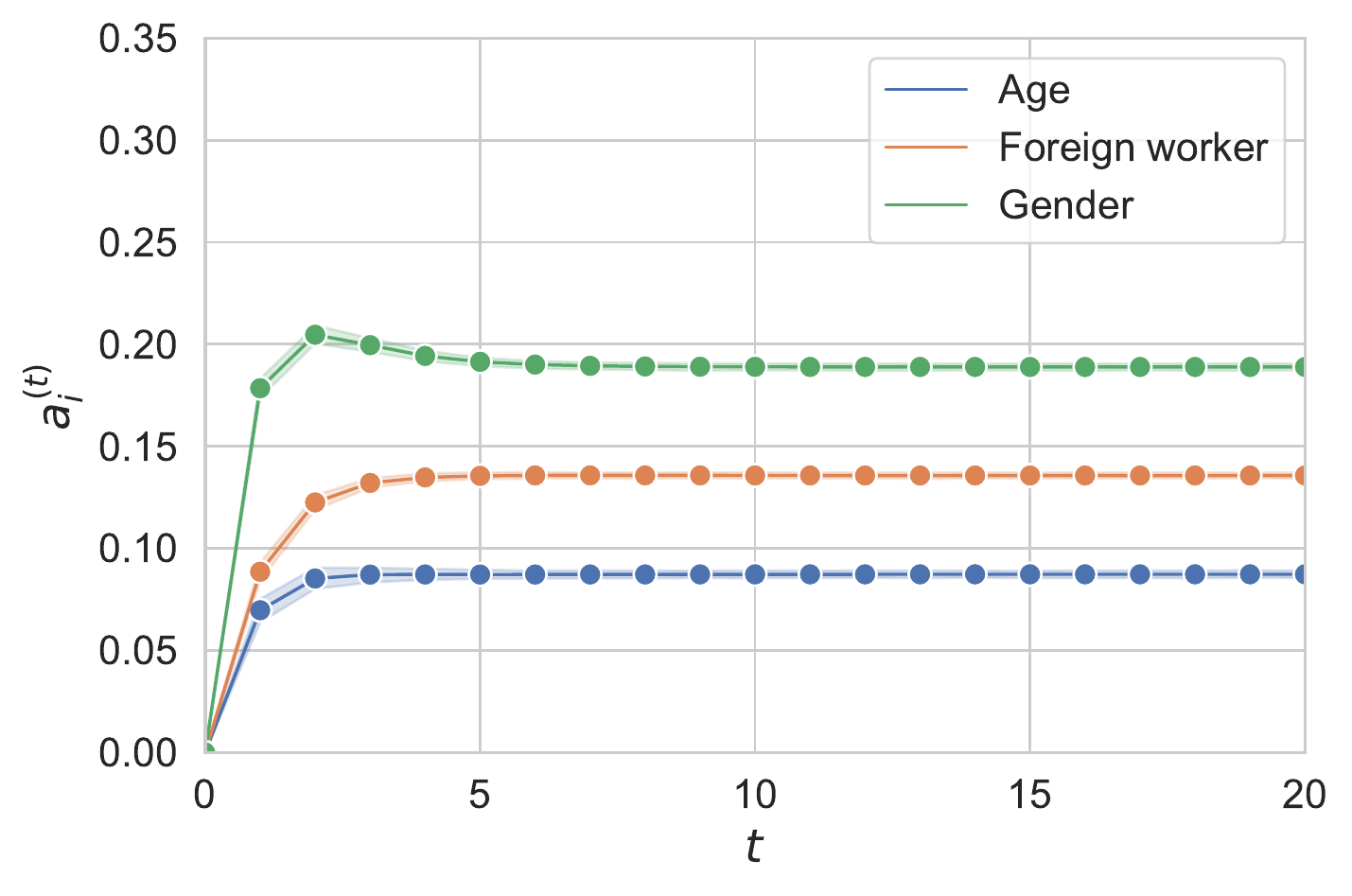}
	\caption{$\phi=0.8$}
	\end{subfigure}
	\begin{subfigure}{0.33\textwidth}
	\center
	\includegraphics[width=\textwidth]{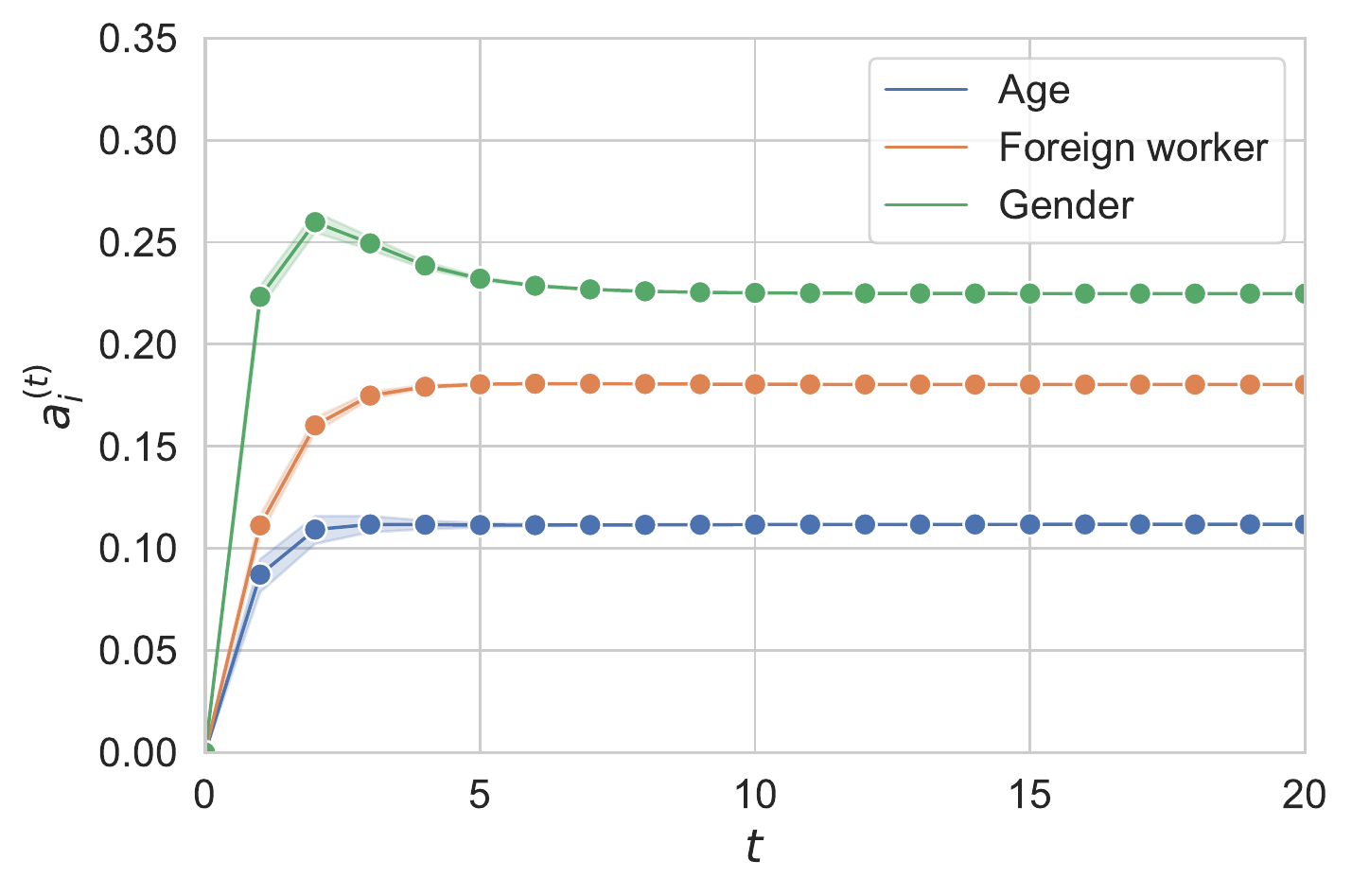}
	\caption{$\phi=1.0$}
	\end{subfigure}
	
	\captionsetup{justification=justified}
	\caption{Activation values of neurons denoting protected features for the third scenario. The unprotected features \textit{housing}, \textit{people liable}, \textit{employment since} and \textit{purpose}, which have the largest correlation with the protected feature \textit{gender}, are randomly initialized when running the simulations.}
\label{fig:phi_var_gender}
\end{figure*}

\begin{figure*}[!htbp]
\center

	\begin{subfigure}{0.33\textwidth}
	\center
	\includegraphics[width=\textwidth]{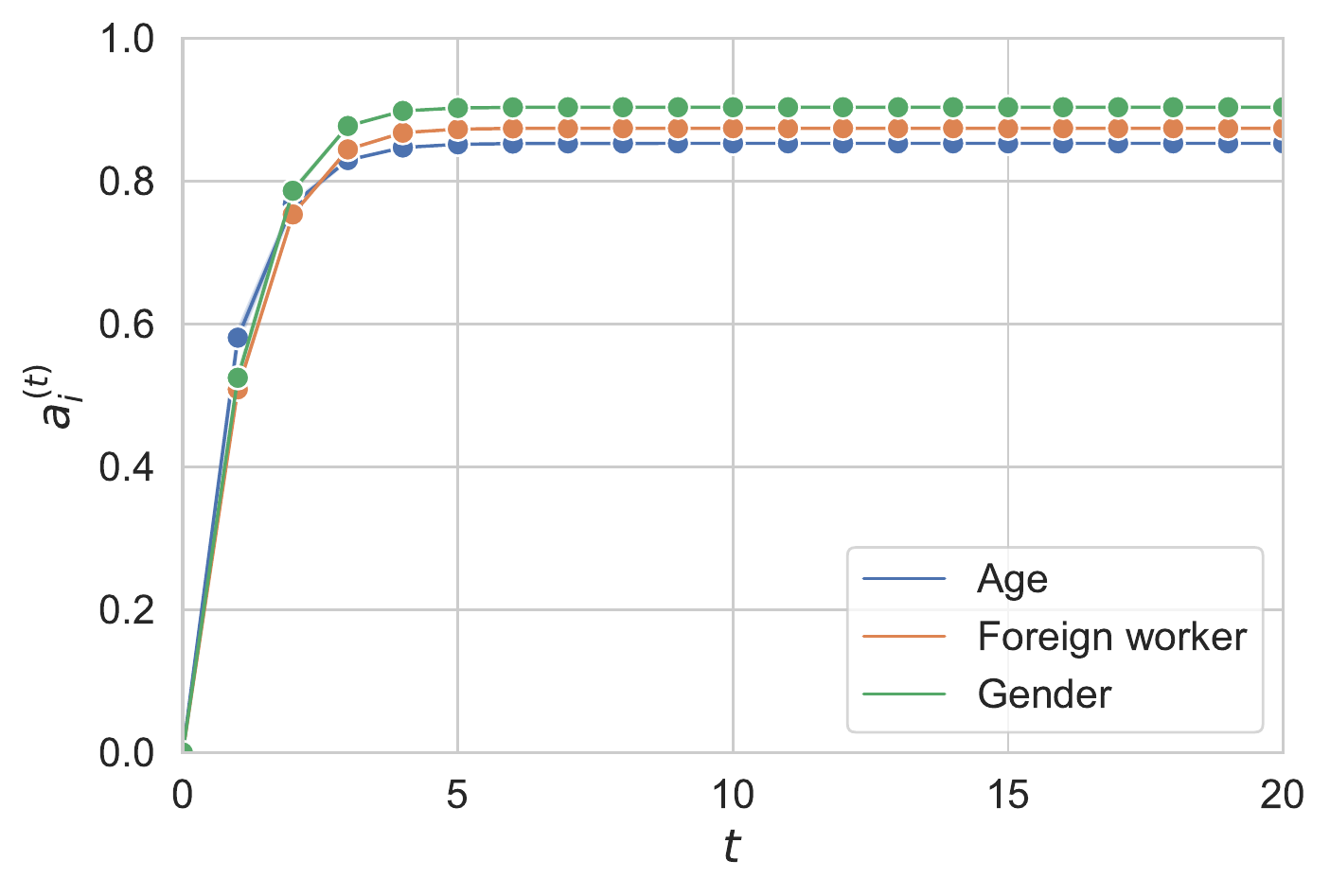}
	\caption{Sigmoid activation function}
	\end{subfigure}
	\begin{subfigure}{0.33\textwidth}
	\center
	\includegraphics[width=\textwidth]{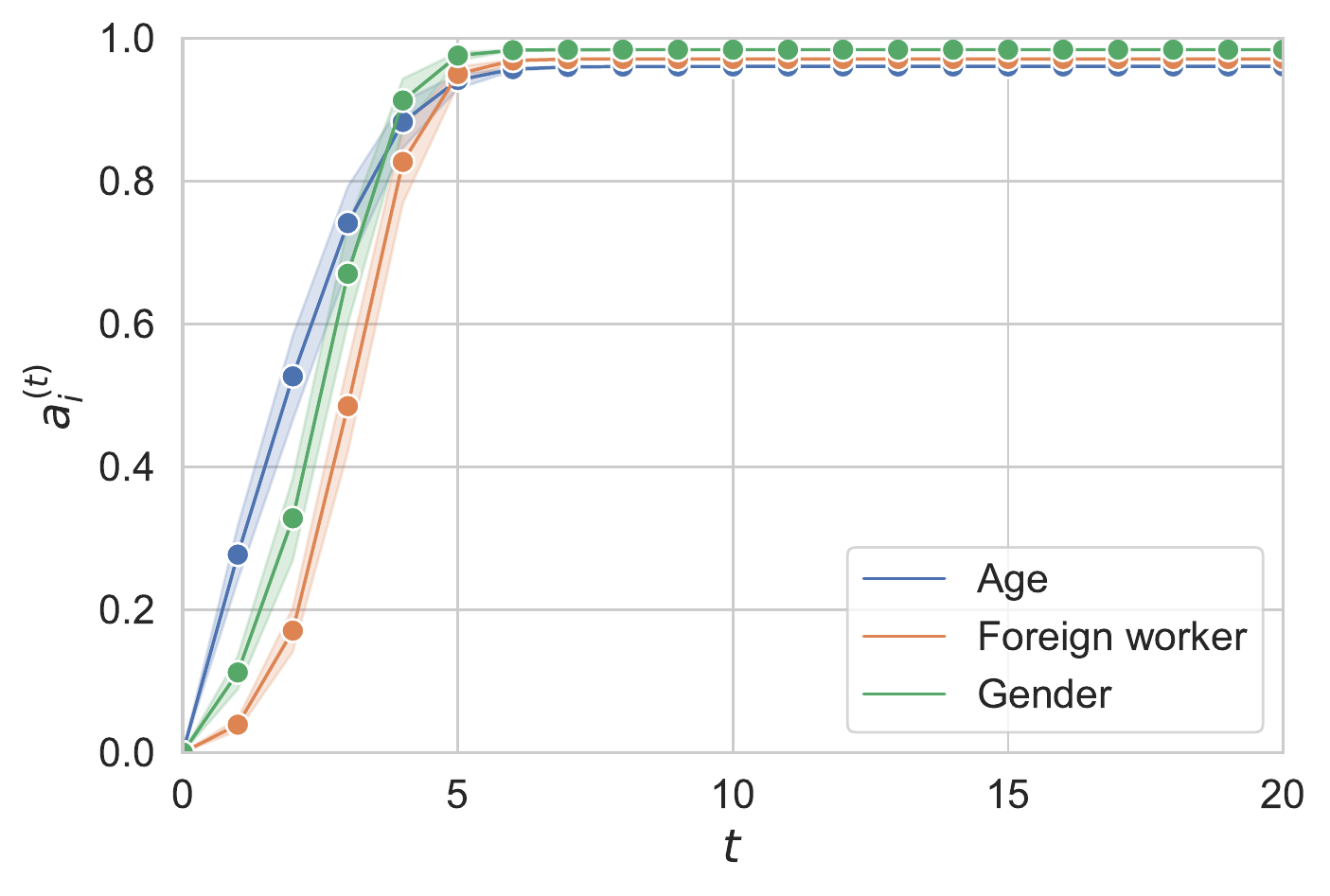}
	\caption{Hyperbolic activation function}
	\end{subfigure}
	\begin{subfigure}{0.33\textwidth}
	\center
	\includegraphics[width=\textwidth]{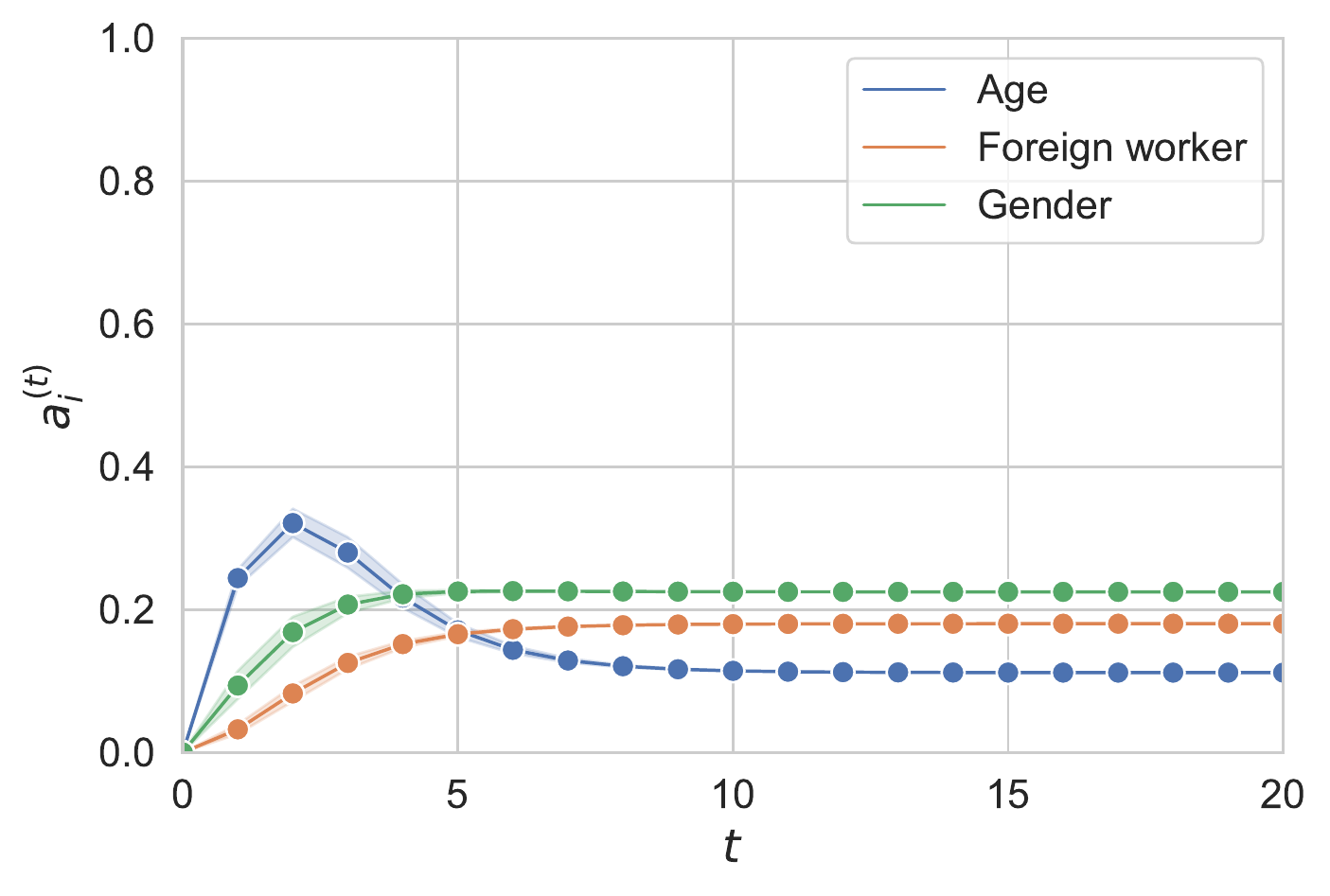}
	\caption{Re-scaled activation function}
	\end{subfigure}
	
	\captionsetup{justification=justified}
	\caption{Activation values of neurons denoting protected features for the first scenario, using different activation functions. Although the activation of \textit{gender} is greater than the activation of \textit{age} or \textit{foreign worker} for all functions, the sigmoid and the hyperbolic tangent functions seem to be less sensitive to this difference. Meanwhile the proposed re-scaled activation function is able to capture the difference without saturating the neurons.}
	
\label{fig:fixed-points}
\end{figure*}

According to our definition of implicit bias, one would expect each scenario to be more biased towards the protected feature that correlates with the unprotected ones that were activated. Let us analyze the simulation results. Figures \ref{fig:phi_var_age}, \ref{fig:phi_var_foreign} and \ref{fig:phi_var_gender} show the activation values of neurons representing protected features for $T=20$ and different $\phi$ values for the three scenarios, respectively. As expected, in the third scenario, the activation values of the neuron denoting \textit{gender} are consistently larger than the ones for \textit{age} or \textit{foreign worker}, when varying the $\phi$ parameter. In the first and second scenarios, however, we see that the model reports more bias towards \textit{gender} than \textit{age} or \textit{foreign worker}, as $\phi$ approaches one. In other words, the bias against \textit{gender} increases as the neurons' activation values are determined by the interaction among the neurons and the correlation patterns encoded in the weight matrix.

The reader can notice that the final state of our FCM-based model matches in all scenarios when $\phi=1$. This means that the map will produce the same output regardless of the initial activation vector, thus indicating the presence of a unique fixed-point attractor with larger bias towards gender. This result is consistent with our previous research devoted to quantifying explicit bias using fuzzy-rough sets \cite{koutsoviti2021bias}. Moreover, these simulations provide experimental evidence about the uniqueness of the fixed-point attractor when the nonlinear component in Equation \eqref{eq:reasoning-new} is entirely suppressed.

Now, let us suppose that we want to build a decision-making model concerning the first scenario in which the bias towards \textit{gender} must be reduced at least 50\% compared to the implicit bias encoded in the original data. Using the quasi nonlinear reasoning rule, we can conclude that we can rely on up to 60\% of the correlation patterns in the data ($\phi=0.6$) when activating the unprotected features \textit{existing credits}, \textit{employment since} and \textit{residence since}. More explicitly, we can see in Figure \ref{fig:phi_var_age}~(c) that the amount of bias concerning gender is about 0.22 when using $\phi=1$. Such an amount goes down to 0.11 (50\%) as shown in Figure \ref{fig:phi_var_age}~(a) when using $\phi=0.6$, which means that the model is based on 60\% of the correlation patterns and 40\% on the initial conditions. This suggests that our proposal allows making fairer decisions and reusing the historical data that might contain relevant pieces of knowledge.

Finally, we compare the differences in the fixed points produced by the FCM model when suppressing the linear component in Equation \eqref{eq:reasoning-new} and using different transfer functions. Figure \ref{fig:fixed-points} shows the results for the second scenario. We selected a single scenario to keep the simulation simple but the same idea can be extrapolated to the remaining scenarios. 

Observe that the neurons' activation values are more similar among them (and closer to one) when using either the sigmoid or hyperbolic tangent function, which would suggest that neurons are in risk of saturation. This situation will worsen if we increase the slope of these transfer functions. Overall, saturated neurons do not allow making reliable decisions since their activation values offer little possibilities to discriminate among the options. In contrast, the activation values produced by our re-scaled transfer function allow for better separation between the decisions while being far from one. The latter indicates that there are other active neurons in that iteration, otherwise their activation values would be closer to one.

\subsection{Comparison with state-of-the-art methods}
\label{sec:sota}

Before diving into the comparison with other state-of-the-art techniques for detecting (implicit) bias, it is important to remark that our model is originally intended to work at the feature level. Similarly to the FRU measure \cite{koutsoviti2021bias,Napoles2021Afuzzyrough}, our proposal quantifies the bias towards interesting protected features. However, the majority of the approaches in the fairness literature \cite{bonchi2017exposing,aif360-oct-2018,zemel2013learning,Feldman2015} do this analysis at a group level, i.e., defining groups for the protected features. These methods usually discretize protected numerical features in two or more groups, where at least one is considered unprivileged. Although, the choice of thresholds for these groups can influence the results, even leading to masking the discrimination \cite{corbettdavies2018measure}.

To make our model comparable with these methods, we replace the FCM's concepts denoting protected features by concepts symbolizing protected groups. For the protected features \textit{age} and \textit{gender}, we use the codification in \cite{bonchi2017exposing} involving four possible values each. The groups for \textit{age} are as follows: people younger than 30 years of age (\textit{age\_le\_30}), people from 30 to 41 years old (\textit{age\_from\_30\_le\_41}), people from 41 to 52 years old (\textit{age\_from\_41\_le\_52}), and people older than 52 years of age (\textit{age\_gt\_52}). The groups for \textit{gender} have extra information on the civil status and categorize females independently of their civil status (\textit{gender\_female}), single males (\textit{gender\_male\_single}), married or widowed males (\textit{gender\_male\_mar\_or\_wid}), and divorced or separated males (\textit{gender\_male\_div\_or\_sep}). For the protected feature \textit{foreign worker}, we will have two groups representing the binary options. Numerical unprotected features will not be discretized to avoid losing information, therefore, the new FCM contains 27 concepts.

\begin{figure}[ht]
    \centering
    \resizebox{\linewidth}{!}{
    \includegraphics{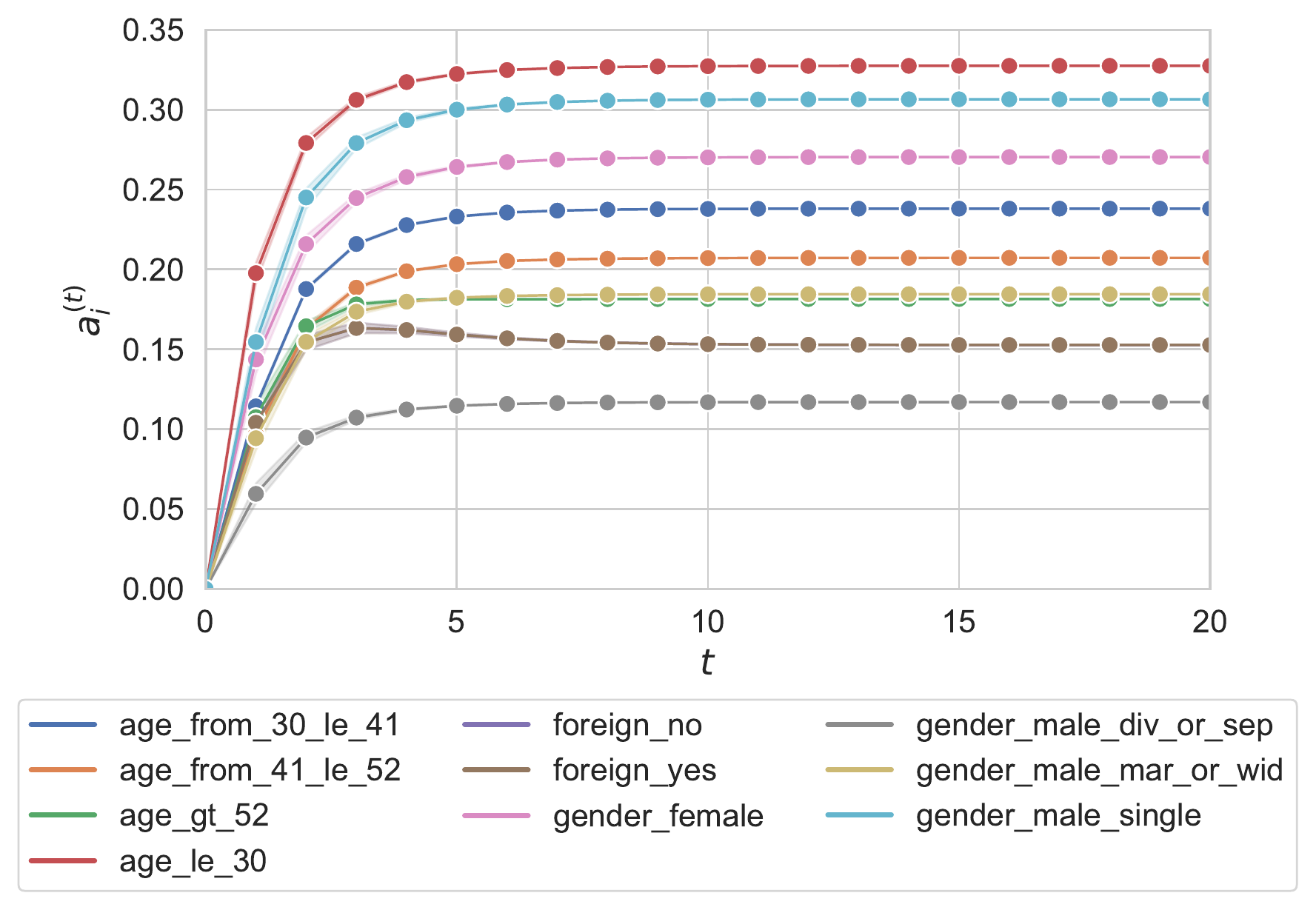}}
    \caption{Activation values for the neural concepts denoting protected groups. The proposed FCM model signals a higher bias against people under 30 years old and single males than the remaining groups.}
    \label{fig:group}
\end{figure}

As for the nonlinearity degree, we choose $\phi=1$ to guarantee the convergence to a unique fixed-point attractor. Nevertheless, we generate 20 initial activation vectors where randomly selected concepts denoting unprotected features are randomly initialized, whereas the remaining ones are set to zero. Figure \ref{fig:group} shows the results of the simulation.

According to our FCM model, the concept with the highest activation value concerns young individuals (\textit{age\_le\_30}). This result agrees with traditional statistical bias detection measures such as statistical parity difference \citep{Dwork2012}, equal opportunity difference \cite{Hardt2016}, average odds difference \cite{Hardt2016}, and disparate impact \cite{Feldman2015}, which signal high bias towards young individuals. Interestingly, when using the codification for \textit{gender} that includes information on the civil status, we obtain that the single males (\textit{gender\_male\_single}) is one of the most discriminated groups, as also concluded by \cite{bonchi2017exposing}. Moreover, our model keeps indicating a strong bias against females and moderate bias towards middle age people. Less bias is detected towards foreign workers, older people, or males other than single. 

It is worth further discussing the divergence between the results reported in \cite{bonchi2017exposing} and the ones obtained with our model for the protected feature age. The mentioned study attributes more bias against people older than 52 years and less discrimination against young people. However, when looking at the distribution of good and bad credits for each of the groups associated with age (see Table \ref{tab:dist}), it is evident that younger applicants are less favored than older ones. More explicitly, the percentage of people younger than 30 years with respect to the total of bad credits (45\%) is significantly larger that the same demographic for the good credit (33\%). In contrast, the distribution of elderly people is very similar for obtaining a bad or a good credit, with 10\% and 11\%, respectively.

\begin{table}[!h]
\centering
\caption{Distribution of good and bad credits per age group for the German credit dataset.}
\begin{tabular}{c|c|c}
\hline
Age Groups       & Bad Credit & Good Credit \\ \hline
le\_30           & 137 (45\%) & 234 (33\%)  \\
from\_30\_le\_41 & 91 (30\%)  & 264 (38\%)  \\
from\_41\_le\_52 & 42 (15\%)  & 127 (18\%)  \\
gt\_52           & 30 (10\%)  & 75 (11\%)   \\ \hline
Total            & 300        & 700         \\ \hline
\end{tabular}
\label{tab:dist}
\end{table}

\smallskip
\smallskip

Before concluding the paper, let us summarize the main features of our proposal. First, the FCM model determines implicit bias in situations where pairwise analysis might lead to misleading results. Second, FCMs are able to handle both numeric and nominal features which minimizes data pre-processing and preserves information in its original state.  Third, we suggest an interpretable way to rely on the data to a certain extent when taking decisions regarding to a sensitive feature. In practice, this feature allows decision makers to complement their data-driven decision making approach with alternative practices. Finally, the introduction of the nonlinear component expressed by the $\phi$ value and the re-scaled transfer function paves the way to perform richer what-if simulations.

\section{Concluding remarks}
\label{sec:remarks}


This paper introduced an FCM-based model to quantify implicit bias in structured classification datasets where features are characterized as protected and unprotected ones. Our model uses correlation patterns as weights connecting the neural concepts since the intuition dictates that implicit bias can be quantified from the correlation between unprotected and protected features. At the same time, we solved relevant theoretical challenges concerning the controllability and convergence of FCM-based simulation models that had remained unsolved. Firstly, we employed a quasi nonlinear neural reasoning mechanism involving a parameter to control nonlinearity in the predictions. Secondly, we proposed a re-scaled transfer function that promotes competition among neurons in each iteration. Finally, we derived analytical conditions ensuring either that the model converges to a unique fixed point or that the unique fixed point does not exist. This means that we obtained an easily-controllable simulation model to either represent invariant situations or perform what-if simulations.

In our numerical simulations, an FCM-based model is built using absolute correlation values computed in a pairwise manner between all features of the German Credit dataset. Since this dataset has been widely examined on the basis of possibly being biased towards the features \textit{age}, \textit{foreign worker} and \textit{gender}, we chose to investigate the flow of information from selected unprotected features to these three sensitive concepts. We built three scenarios: we used unprotected features that strongly correlate with each sensitive feature as activation neurons. The resulting activation values of neurons denoting a protected feature are conceptually equivalent to its relevance inside the system and are used to estimate the implicit bias flowing from the activated features towards them. In all scenarios, results show that activation values of the neuron representing \textit{gender} are larger than the ones for \textit{age} or \textit{foreign worker} when $\phi = 1$ regardless of the initial activation vector, indicating that the system converged to a unique fixed point attractor. This denotes that the dataset is more biased against \textit{gender} than \textit{age} or \textit{foreign worker}, a results consistent with our previous models using entirely different approaches  \cite{koutsoviti2021bias, Napoles2021Afuzzyrough}. A second finding arises when modeling bias that is implicitly reflected on \textit{age} from highly correlated unprotected features: \textit{age}'s activation values gradually become smaller than those of \textit{gender} as we increasingly allow the neurons to interact without any constraints. This implies that one needs to examine different $\phi$ levels in order to select the optimal level.

Actually, we can formulate a constrained optimization problem to find out the maximal nonlinearity amount and activation values for given protected features such that we minimize the bias towards the protected features. The constraints can be used to ensure that the relevance of interesting features does not fall below a given threshold or that a desired order relationship among features is preserved. Such a research line will be the target of our future research efforts.




\bibliographystyle{elsarticle-harv}

\end{document}